\documentclass[10pt]{article}
\usepackage{scoai-report}
\usepackage{xurl}
\usepackage[section]{placeins}
\usepackage{float}
\usepackage{needspace}
\renewcommand{\SCOAIBrand}{ScoAI}
\renewcommand{\SCOAISeries}{SC + OR + AI}
\renewcommand{\SCOAITeamEN}{Supply Chain Tech Team Y, JD.com, China}
\renewcommand{\SCOAIDomain}{Forecasting}
\renewcommand{\SCOAITitle}{\fontsize{18}{22}\selectfont TimeHF 2.0: Multimodal, Multivariate,\\and Multi-Region Collaborative Forecasting\\for Large-Scale Retail Supply Chains}
\renewcommand{\SCOAIShortTitle}{TimeHF 2.0}
\renewcommand{\SCOAIAuthors}{%
\mbox{Zhongxuan Han\textsuperscript{1}},
\mbox{Zhengxin Shi\textsuperscript{1}},
\mbox{Kunxiu Chen\textsuperscript{1}},
\mbox{Xiaolin Huang\textsuperscript{1}},
\mbox{Yulin Huang\textsuperscript{1}},
\mbox{Anni Zhang\textsuperscript{1}},
\mbox{Xiaoming Lin\textsuperscript{1}},
\mbox{Zhengyu Chen\textsuperscript{1}},
\mbox{Xue Yu\textsuperscript{1}},
\mbox{Yang Liu\textsuperscript{1}},
\mbox{Yuanyuan Yang\textsuperscript{1}},
\mbox{Yangyang Zhang\textsuperscript{1}},
\mbox{Junheng Huang\textsuperscript{1}},
\mbox{Dongyang Geng\textsuperscript{1}},
\mbox{Ben Zhang\textsuperscript{1}},
\mbox{Xin Wang\textsuperscript{1}},
\mbox{Jiaxin Wang\textsuperscript{1}},
\mbox{Haowei Zhou\textsuperscript{1}},
\mbox{Qiqi Yan\textsuperscript{1}},
\mbox{Jiaheng Yin\textsuperscript{1}},
\mbox{Jintao Xu\textsuperscript{1}},
\mbox{Zijiang Liu\textsuperscript{1}},
\mbox{Changfei Zhan\textsuperscript{1}},
\mbox{Zhe Chen\textsuperscript{1}},
\mbox{Yutong Pan\textsuperscript{1}},
\mbox{Jinqiu Deng\textsuperscript{1}},
\mbox{Shaomin Li\textsuperscript{1}},
\mbox{Ningxuan Kang\textsuperscript{1}},
\mbox{Jianshen Zhang\textsuperscript{1}},
\mbox{Yongzhi Qi\textsuperscript{1,*}},
\mbox{Wei Qi\textsuperscript{2}},
\mbox{Dazhou Lei\textsuperscript{3}},
\mbox{Zuo-Jun Max Shen\textsuperscript{4}}%
}
\renewcommand{\SCOAIPDFAuthors}{Zhongxuan Han, Zhengxin Shi, Kunxiu Chen, Xiaolin Huang, Yulin Huang, Anni Zhang, Xiaoming Lin, Zhengyu Chen, Xue Yu, Yang Liu, Yuanyuan Yang, Yangyang Zhang, Junheng Huang, Dongyang Geng, Ben Zhang, Xin Wang, Jiaxin Wang, Haowei Zhou, Qiqi Yan, Jiaheng Yin, Jintao Xu, Zijiang Liu, Changfei Zhan, Zhe Chen, Yutong Pan, Jinqiu Deng, Shaomin Li, Ningxuan Kang, Jianshen Zhang, Yongzhi Qi, Wei Qi, Dazhou Lei, Zuo-Jun Max Shen}
\renewcommand{\SCOAIAffiliations}{%
\textsuperscript{1}Supply Chain Tech Team Y, JD.com, China\\
\textsuperscript{2}Tsinghua University, China\\
\textsuperscript{3}Beijing Jiaotong University, China\\
\textsuperscript{4}The University of Hong Kong, China%
}
\renewcommand{\SCOAIAuthorNote}{\textsuperscript{*}Corresponding author: Yongzhi Qi.}
\renewcommand{\SCOAIKeywords}{Supply-chain forecasting; business knowledge; multivariate streams; regional collaboration; mixture of experts}

\newcommand{\TimeHFPlainTitle}{TimeHF 2.0: Multimodal, Multivariate, and Multi-Region Collaborative Forecasting for Large-Scale Retail Supply Chains}

\newcommand{\model}{TimeHF 2.0}

\newcommand{\ind}{\mathbf{1}}
\DeclareMathOperator{\asinh}{asinh}

\DeclareMathOperator{\RMS}{RMS}
\DeclareMathOperator{\LN}{Norm}
\definecolor{rivetblue}{HTML}{234D68}
\definecolor{rivetink}{HTML}{203744}
\definecolor{rivetregion}{HTML}{3D6392}
\definecolor{rivetteal}{HTML}{247D71}
\definecolor{rivetpurple}{HTML}{765394}
\definecolor{rivetgold}{HTML}{AA782F}

\usepackage{pgfplots,pgfplotstable}
\pgfplotsset{compat=1.18}
\definecolor{EvalFocus}{HTML}{D1642B}
\definecolor{EvalGray}{HTML}{87939C}
\definecolor{EvalNational}{HTML}{234D68}
\definecolor{EvalPeerB}{HTML}{688A9B}
\definecolor{EvalPeerC}{HTML}{A4ADB5}

\newcommand{\Estimate}[3]{\ifdefempty{#1}{\phantom{+00.0}}{\ifdefempty{#2}{#1}{\shortstack{#1\\[-1pt]{\scriptsize[#2,\,#3]}}}}}

\newcommand{\Est}[1]{\begingroup\expandafter\let\expandafter\ResultValue\csname R#1\endcsname\expandafter\let\expandafter\ResultLower\csname R#1Low\endcsname\expandafter\let\expandafter\ResultUpper\csname R#1High\endcsname\Estimate{\ResultValue}{\ResultLower}{\ResultUpper}\endgroup}
\newtcolorbox{evalcase}[2]{scoai base,breakable=false,colback=white,colframe=SCOAIRule,colbacktitle=SCOAIPale,coltitle=SCOAISlate,title={#1\enspace /\enspace #2},left=7pt,right=7pt,top=4pt,bottom=4pt,before skip=7pt,after skip=7pt}
\newtcolorbox{knowledgebox}{boxrule=0pt,leftrule=1.6pt,colframe=EvalFocus,colback=SCOAIWarm,arc=0pt,left=6pt,right=6pt,top=4pt,bottom=4pt,before skip=4pt,after skip=4pt,fontupper=\footnotesize\raggedright}

\pgfplotsset{eval axis/.style={width=.98\linewidth,height=4.1cm,scale only axis=false,xmin=-\CasePlotHistoryDays,xmax=\CasePlotHorizon,ymin=0,ymax=\CasePlotUpper,axis lines=left,axis line style={SCOAIMuted!55},tick style={SCOAIMuted!55},tick label style={font=\scriptsize,color=SCOAIMuted},label style={font=\scriptsize,color=SCOAIMuted},xtick={-\CasePlotHistoryDays,0,\CasePlotHorizon},xticklabels={$T-L$,$T$,$T+H$},ytick=\empty,ylabel={Indexed sales},grid=none,unbounded coords=jump,filter discard warning=false,legend style={font=\scriptsize,draw=none,fill=none,at={(.5,-.23)},anchor=north,legend columns=3,column sep=5pt},clip=false}}
\newcommand{\ReadCaseTable}[1]{\pgfplotstableread[col sep=comma]{#1}\CaseTable\pgfplotstablegetrowsof{\CaseTable}\pgfmathtruncatemacro{\CaseRows}{\pgfplotsretval}}
\newcommand{\KnowledgeSalesLabel}{Observed sales}
\newcommand{\KnowledgePlot}[2][]{\ReadCaseTable{data/#2.csv}\begin{tikzpicture}\begin{axis}[eval axis,#1]
\draw[densely dashed,SCOAIMuted!70] (axis cs:0,0)--(axis cs:0,\pgfkeysvalueof{/pgfplots/ymax});
\ifnum\CaseRows>0
\addplot[SCOAIInk,line width=.75pt] table[x=lead,y=observed]{\CaseTable};
\addplot[EvalGray,line width=.85pt,dash pattern=on 1pt off 1pt on 3pt off 1pt] table[x=lead,y=baseline]{\CaseTable};
\addplot[EvalFocus,line width=1.15pt,dashed] table[x=lead,y=conditioned]{\CaseTable};
\else
\addlegendimage{SCOAIInk,line width=.75pt}\addlegendimage{EvalGray,line width=.85pt,dash pattern=on 1pt off 1pt on 3pt off 1pt}\addlegendimage{EvalFocus,line width=1.15pt,dashed}
\fi
\legend{\KnowledgeSalesLabel,Original context,Updated context}
\end{axis}\end{tikzpicture}\par\vspace{3pt}}
\newcommand{\RegionalObservedLabel}{Observed sales}
\newcommand{\RegionalPlot}[2][]{\ReadCaseTable{data/#2.csv}\begin{tikzpicture}
\begin{axis}[eval axis,name=overview,width=.94\linewidth,height=3.35cm,legend style={font=\scriptsize,draw=none,at={(.5,1.48)},anchor=south,legend columns=4,column sep=5pt},title={\scriptsize A\quad National and DC trajectories},title style={at={(.5,1.03)},anchor=south},#1]
\draw[densely dashed,SCOAIMuted!70] (axis cs:0,0)--(axis cs:0,\CasePlotUpper);
\ifnum\CaseRows>0
\addplot[EvalNational,line width=.6pt,forget plot] table[x=lead,y=national_history]{\CaseTable};\addplot[EvalNational,line width=.6pt,dashed] table[x=lead,y=national_forecast]{\CaseTable};
\addplot[EvalPeerB,line width=.55pt,forget plot] table[x=lead,y=peer_b_history]{\CaseTable};\addplot[EvalPeerB,line width=.55pt,dashed] table[x=lead,y=peer_b_forecast]{\CaseTable};
\addplot[EvalPeerC,line width=.55pt,forget plot] table[x=lead,y=peer_c_history]{\CaseTable};\addplot[EvalPeerC,line width=.55pt,dashed] table[x=lead,y=peer_c_forecast]{\CaseTable};
\addplot[EvalFocus,line width=1.15pt,forget plot] table[x=lead,y=focal_history]{\CaseTable};\addplot[EvalFocus,line width=1.15pt,dashed] table[x=lead,y=focal_forecast]{\CaseTable};
\else
\addlegendimage{EvalNational,line width=.6pt}\addlegendimage{EvalPeerB,line width=.55pt}\addlegendimage{EvalPeerC,line width=.55pt}\addlegendimage{EvalFocus,line width=1.15pt}
\fi
\legend{National,DC-B,DC-C,Focal DC-A}
\end{axis}
\begin{axis}[eval axis,at={(overview.south west)},anchor=north west,yshift=-1.15cm,width=.94\linewidth,height=3.05cm,title={\scriptsize\color{EvalFocus} B\quad Focal DC-A forecast check},legend style={font=\scriptsize,draw=none,at={(.5,-.24)},anchor=north,legend columns=3,column sep=5pt},#1]
\draw[densely dashed,SCOAIMuted!70] (axis cs:0,0)--(axis cs:0,\CasePlotUpper);
\ifnum\CaseRows>0
\addplot[SCOAIInk,line width=.7pt,forget plot] table[x=lead,y=focal_history]{\CaseTable};\addplot[SCOAIInk,line width=.7pt] table[x=lead,y=focal_observed]{\CaseTable};
\addplot[EvalGray,line width=.8pt,dash pattern=on 1pt off 1pt on 3pt off 1pt] table[x=lead,y=independent]{\CaseTable};
\addplot[EvalFocus,line width=1.1pt,dashed] table[x=lead,y=focal_forecast]{\CaseTable};
\else
\addlegendimage{SCOAIInk,line width=.7pt}\addlegendimage{EvalGray,line width=.8pt,dash pattern=on 1pt off 1pt on 3pt off 1pt}\addlegendimage{EvalFocus,line width=1.1pt,dashed}
\fi
\legend{\RegionalObservedLabel,No regional messages,Joint forecast}
\end{axis}
\end{tikzpicture}\par\vspace{5pt}}

\definecolor{CaseReference}{HTML}{278A78}
\definecolor{CaseContext}{HTML}{78858D}
\renewtcolorbox{evalcase}[2]{scoai base,breakable=false,colback=white,
  colframe=SCOAIRule,colbacktitle=SCOAIPale,coltitle=SCOAISlate,
  title={#1\enspace /\enspace #2},left=7pt,right=7pt,top=4pt,bottom=4pt,
  before skip=3pt,after skip=3pt}
\pgfplotsset{case axis/.style={
  axis x line*=bottom,axis y line*=left,
  axis line style={SCOAIMuted!45},tick style={SCOAIMuted!45},
  tick label style={font=\scriptsize,color=SCOAIMuted},
  label style={font=\scriptsize,color=SCOAIMuted},
  title style={font=\footnotesize,color=SCOAISlate},
  scaled y ticks=false,/pgf/number format/1000 sep={,},
  ymin=0,grid=none,unbounded coords=jump,
  filter discard warning=false,clip=true,clip mode=individual}}

\newcommand{\NationalOneDailyGain}{13.84}
\newcommand{\NationalTwoDailyGain}{18.25}
\newcommand{\NationalOneWindowGain}{13.27}
\newcommand{\NationalTwoWindowGain}{17.94}
\newcommand{\DCOneDailyGain}{12.02}
\newcommand{\DCNoRegionDailyGain}{17.88}
\newcommand{\DCNoCovariateDailyGain}{17.34}
\newcommand{\DCNoMultimodalDailyGain}{18.02}
\newcommand{\DCNoCumulativeDailyGain}{17.71}
\newcommand{\DCFullDailyGain}{19.44}
\newcommand{\DCOneWindowGain}{11.56}
\newcommand{\DCNoRegionWindowGain}{16.96}
\newcommand{\DCNoCovariateWindowGain}{16.03}
\newcommand{\DCNoMultimodalWindowGain}{17.65}
\newcommand{\DCNoCumulativeWindowGain}{16.50}
\newcommand{\DCFullWindowGain}{18.99}
\newcommand{\LossMSENational}{15.02}
\newcommand{\LossMSEDC}{17.34}
\newcommand{\LossMAENational}{15.86}
\newcommand{\LossMAEDC}{17.75}
\newcommand{\LossHuberNational}{16.34}
\newcommand{\LossHuberDC}{17.99}
\newcommand{\LossBinsNational}{17.12}
\newcommand{\LossBinsDC}{18.54}
\newcommand{\BestOneNational}{16.23}
\newcommand{\BestOneDC}{13.44}
\newcommand{\BestTwoNational}{19.52}
\newcommand{\BestTwoDC}{20.11}
\newcommand{\SeasonOneNational}{12.16}
\newcommand{\SeasonOneDC}{9.45}
\newcommand{\SeasonTwoNational}{17.64}
\newcommand{\SeasonTwoDC}{24.52}
\newcommand{\TailOneNational}{15.77}
\newcommand{\TailOneDC}{12.64}
\newcommand{\TailTwoNational}{20.15}
\newcommand{\TailTwoDC}{28.66}
\newcommand{\NewOneNational}{3.01}
\newcommand{\NewOneDC}{2.19}
\newcommand{\NewTwoNational}{9.28}
\newcommand{\NewTwoDC}{11.25}

\newcommand{\CasePlotHistoryDays}{30}
\newcommand{\CasePlotHorizon}{28}
\newcommand{\CasePlotUpper}{1000}

\begin{document}
\brokenpenalty=10000
\hypersetup{pdftitle={\TimeHFPlainTitle},pdfauthor={\SCOAIPDFAuthors}}
\SCOAIMakeTitle
\begin{SCOAIAbstract}
Large-scale retail supply chains need sales forecasts for specified products, future dates or planning windows, and regions under the conditions available at the forecast origin. Regional sparsity, changing environments, volatile sales, and unfamiliar business situations make these dimensions difficult to model together. We introduce \model{}, a multimodal, multivariate, and multi-region collaborative forecasting framework that makes flexibly expressed business knowledge usable in numerical prediction. A coverage-aware directed encoder integrates regional sales and applicable covariates. A semantic alignment module transfers language-model hidden representations of natural-language plans and domain experience into target-specific forecasting context. Learning against sales observations connects these meanings to numerical evidence, allowing business users to supply experience, future plans, and changing circumstances without defining a new numerical feature for each description. Text-conditioned future queries and shared/routed experts adapt prediction to the supplied context. An ordered-distribution head allocates fine resolution to routine fluctuations and coarse resolution to extreme sales movements, while daily/window learning connects forecasts to cumulative planning quantities. Online parallel forecasting at JD yields relative accuracy improvements of 18.25\% and 17.94\% for national daily and planning-window sales, and 19.44\% and 18.99\% for DC daily and planning-window sales, over the original production deep learning baseline. Component comparisons, loss-design results, and product-type analyses examine the roles of regional information, covariates, business knowledge, context-dependent processing, and robust planning-oriented learning.

\end{SCOAIAbstract}
\clubpenalty=10000
\widowpenalty=10000
\displaywidowpenalty=10000
\section{Introduction}
\label{sec:introduction}

Large-scale retail supply chains coordinate procurement, inventory placement, replenishment, and fulfillment across products, locations, and planning horizons~\citep{hu2024supercharged,wang2022digitaltwin}. These decisions need sales estimates that specify how much to source, where stock will be needed, and when to replenish it. The ScoAI perspective places learned estimates within this operational setting, where costs, resource limits, and business rules determine how predictions are used~\citep{liu2026scoai}. Forecasting supplies the common quantity: sales for a specified product, future period, and region under the business conditions known when the decision is made.

At JD, forecasting spans tens of millions of products and hundreds of distribution centers (DCs), with horizons extending multiple weeks. Daily operations require over one hundred million forecast outputs.

\begin{figure}[t]
\centering
\includegraphics[width=0.98\linewidth]{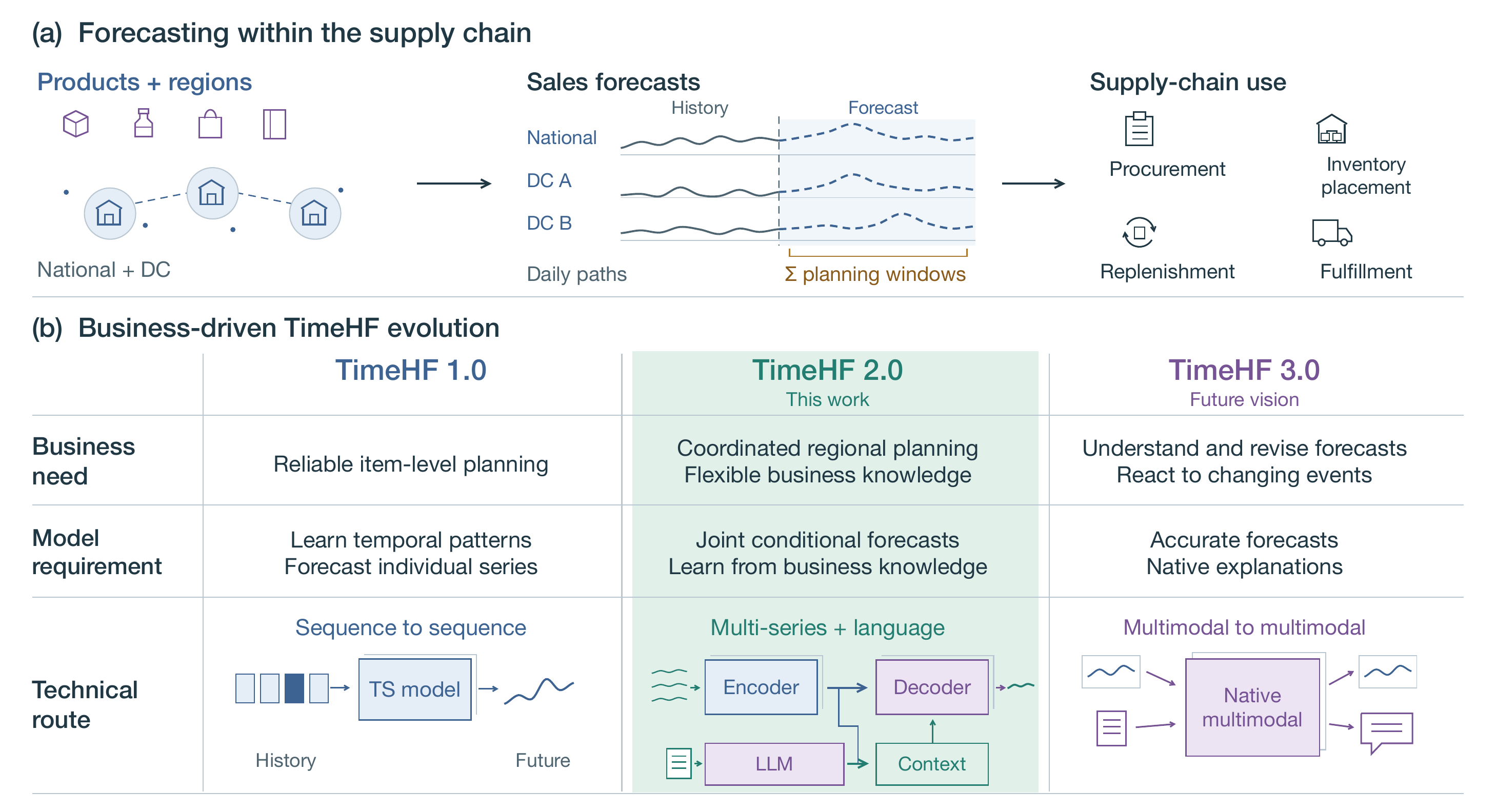}
\caption{\textbf{Supply-chain forecasting context and the TimeHF roadmap.} \textbf{(a)} Solid histories and dashed future trajectories illustrate national/DC forecasts; daily paths and window totals support procurement, inventory placement, replenishment, and fulfillment. The vertical line marks the forecast origin. \textbf{(b)} Each generation connects a business need to required capabilities and a technical route. TimeHF~2.0 is the present framework, connecting regional numerical evidence with language-conditioned forecasting. TimeHF~3.0 envisions native multimodal inputs and outputs, combining forecast accuracy with native explanations. The TimeHF~2.0 route depicts multiple numerical inputs and a single target sales trajectory; the model is applied to national and DC targets. Routes and trajectories are schematic; the first generation's sequence-to-sequence label describes its input/output mapping.}
\label{fig:timehf_roadmap}
\end{figure}

\Needspace{5\baselineskip}
National forecasts inform procurement and supply planning, while DC forecasts inform inventory placement and local replenishment; daily trajectories and cumulative quantities serve different operational windows. Figure~\ref{fig:timehf_roadmap} relates this joint product--time--region--condition forecasting need to the TimeHF roadmap.

JD's TimeHF family follows the business-driven roadmap in Figure~\ref{fig:timehf_roadmap}. TimeHF~1.0~\citep{qi2025timehf} supports reliable item-level planning through temporal modeling and human-feedback-guided optimization, mapping historical sequences to future sales. Coordinating procurement with regional inventory and replenishment requires joint, context-aware forecasts and a flexible way to supply plans and domain experience when comparable history is scarce. TimeHF~2.0 connects a multi-series encoder--decoder with language-model representations through semantic alignment, bringing business knowledge, regional evidence, and structured conditions into the same prediction process.

Changing events and plans also create a need to understand and revise forecasts. The future TimeHF~3.0 direction is native multimodal-to-multimodal modeling, bringing heterogeneous inputs, accurate forecasts, and native explanations into a common interface. This report focuses on TimeHF~2.0; the third stage remains a future vision.

\begin{technicalproblem}[TimeHF~2.0]
For a specified \textbf{product}, specified \textbf{future dates or planning windows}, and a specified \textbf{region} (national or DC), under the applicable \textbf{covariates and business context} known at the forecast origin, \textbf{how many units will sell}?
\end{technicalproblem}

\begin{figure}[t]
\centering
\includegraphics[width=0.98\linewidth]{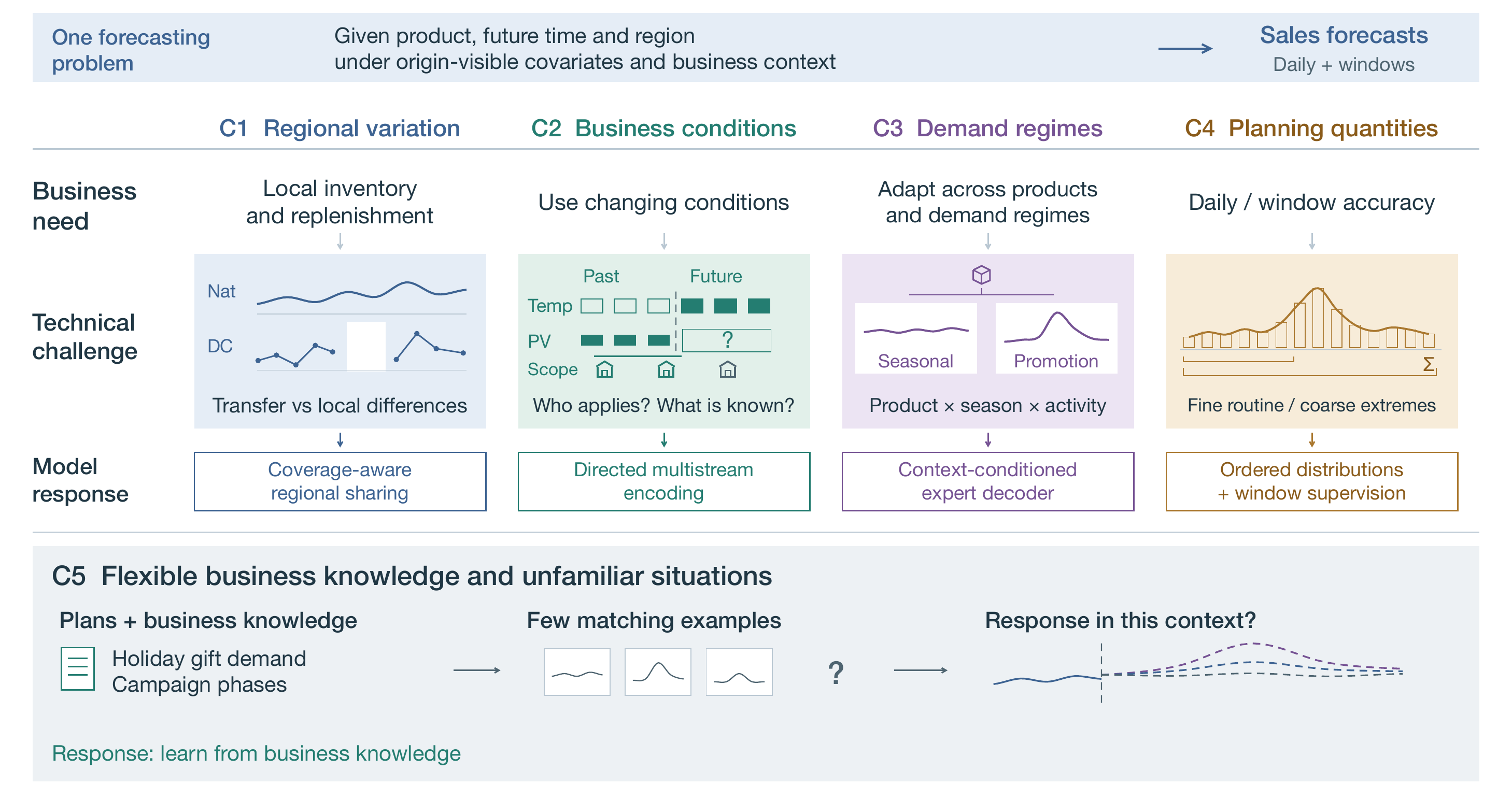}
\caption{\textbf{Business situations and forecasting challenges in TimeHF~2.0.} The top strip states the shared product--time--region forecasting task under available covariate and business context. C1--C4 illustrate regional histories, condition scope and availability, product types and seasonal/activity regimes, and daily/window quantities, including fine routine and coarse extreme variation. C5 connects user-expressed plans and domain experience with unfamiliar situations and scarce comparable evidence. Response labels summarize the corresponding components in Section~\ref{sec:method}. Temp denotes an origin-visible temperature forecast. PV denotes observed page views. Conditions, histories, and response paths are schematic examples, not measured forecasts.}
\label{fig:business_challenges}
\end{figure}

TimeHF~2.0 treats these coordinates as one forecasting problem. Observed sales histories, structured covariates, and flexible business descriptions jointly condition regional daily forecasts and cumulative planning quantities. Five connected challenges arise from this problem (Figure~\ref{fig:business_challenges}).

\Needspace{3\baselineskip}
\noindent\textbf{C1. Regional heterogeneity and limited local histories.}
Inventory placement and local replenishment require useful DC forecasts even when local sales are sparse or recently observed. National and peer-DC histories can provide additional evidence, but regional seasonality, promotion responses, and service footprints differ. The technical challenge is to select and transfer relevant regional information without forcing local sales into a fixed national proportion or transferring unrelated patterns. Sharing must account for the markets served and preserve each target's regional behavior.

\noindent\textbf{C2. Business conditions with different scope and availability.}
Planning relies on changing conditions such as marketing arrangements, regional weather, and page views (PV). Their applicable targets and forecast-time availability differ: an activity schedule may already be confirmed, a weather forecast issued by the origin may cover particular regions and dates, and observed PV can provide only past values. Shared conditions can also produce different regional responses. The technical challenge is to model each condition over time and connect it to the appropriate targets while preserving its role, scope, and availability at the forecast origin.

\noindent\textbf{C3. Heterogeneous products and changing demand regimes.}
A common forecasting service must accommodate stable, seasonal, intermittent, and promotion-driven products across many regions. Product categories, seasonal demand, campaign phases, and regional markets can call for different forecasting transformations; a product can also move between these regimes as its conditions change. The technical challenge is to share common patterns across the retail population while selecting suitable processing for the current product and demand regime, with computational cost suitable for large-scale daily operation.

\noindent\textbf{C4. Daily precision and planning-window quantities.}
Operational decisions require both the timing of sales and the quantities accumulated over procurement or replenishment windows. Retail sales combine routine fluctuations, learnable promotion-driven lifts, and isolated transaction peaks without matching origin-visible covariates. The technical challenge is to learn ordinary variation with useful numerical precision while representing extreme movements at a coarser trend level, so unstable amplitudes do not dominate fitting. Persistent daily errors can also accumulate into incorrect planning quantities. Both daily and cumulative accuracy must remain useful when recent histories are incomplete.

\noindent\textbf{C5. Flexible business knowledge and unfamiliar situations.}
Business teams can describe campaign phases, consumption occasions, and practical knowledge of how categories respond to local conditions in natural language. Such knowledge may become available before comparable product--region situations appear in the historical data, and its relevance depends on the product, place, and future timing. The challenge is to let the forecasting model interpret this flexibly supplied knowledge, relate it to numerical evidence, and learn how it should condition the forecast. This need spans regional references, covariate interpretation, product behavior, and the timing or magnitude of sales.

Existing methods provide foundations for these capabilities. Shared forecasting models use related histories, attributes, and known future inputs~\citep{salinas2017deepar,lim2021tft}; large-scale and multivariate models extend representation learning across datasets, targets, and covariates~\citep{das2024timesfm,woo2024moirai,khwaja2026toto2,arango2025chronosx,ansari2025chronos2}. Sparse experts and hierarchical formulations address specialization and relationships among outputs~\citep{liu2024moirai,rangapuram2021hierarchical}. The remaining design problem in this setting is how to organize regional evidence, structured conditions, and business meaning into the same forecasting computation and connect its outputs to planning quantities. Pretrained language representations provide a semantic starting point, while user descriptions supply plans and task-specific experience. Forecast supervision must connect these inputs to the numerical evidence, so that business knowledge can be learned as a useful prediction condition rather than remaining a separate narrative.

The architecture follows three connected stages. A coverage-aware directed encoder integrates national/DC histories and applicable covariates, addressing C1 and C2. Semantic alignment then connects language-model representations of users' plans and experience to target-specific numerical memory, addressing C5. The decoder uses this context in future queries and shared/routed expert processing to adapt across product and demand regimes in C3. An ordered-distribution head and daily/window supervision connect these states to sales trajectories and planning quantities, addressing C4.

The report develops three connected contributions:
\begin{itemize}
  \item A joint forecasting representation that connects regional histories and business conditions through coverage-aware sharing and directed information flow.
  \item A business-knowledge conditioning mechanism that transfers language-model representations of flexible plans and experience into the forecasting space, aligns them with numerical evidence, and learns their use through sales supervision.
  \item A prediction and learning formulation that connects context-dependent specialization to daily sales distributions and cumulative planning quantities at national and DC levels.
\end{itemize}

Online parallel forecasting provides the evaluation setting. Relative to the original production deep learning baseline, \model{} improves national daily and planning-window accuracy by 18.25\% and 17.94\%, and DC daily and planning-window accuracy by 19.44\% and 18.99\%, respectively. Component, loss-design, and product-type comparisons connect the results to regional evidence, structured and textual conditions, robust treatment of volatility, and context-dependent processing. Section~\ref{sec:related} positions the design; Section~\ref{sec:method} connects its components; Section~\ref{sec:experiments} evaluates the resulting daily trajectories and operational planning quantities.

\section{Related Work}
\label{sec:related}
The challenges in the Introduction connect four research strands: shared temporal learning, language conditioning, regional collaboration, and conditional prediction with planning quantities. We review the evidence each strand makes available and the relationships it imposes on outputs, then use these distinctions to motivate the forecasting computation developed in the Method section.

\subsection{Shared and Multivariate Forecasting}
Global forecasting learns across targets through recurrent, state-space, and residual-basis models~\citep{salinas2017deepar,rangapuram2018deepstate,oreshkin2020nbeats}. Retail-specific research addresses SKU-level seasonality~\citep{lin2021seasonal} and new-product life cycles conditioned on attributes and promotions~\citep{lei2023lifecycle}. Temporal Fusion Transformers integrate static attributes and known future inputs~\citep{lim2021tft}, while TimeHF connects large-scale forecasting with human feedback in supply chains~\citep{qi2025timehf}. Transformer research further distinguishes channel-independent patch learning~\citep{nie2023patchtst} from explicit cross-dimension and variate-centric modeling~\citep{zhang2023crossformer,liu2024itransformer}.

Pretrained models extend transfer across datasets using patched decoders, quantized observations, or flexible multivariate attention~\citep{das2024timesfm,ansari2024chronos,woo2024moirai}. ChronosX adds exogenous-variable conditioning, and Chronos-2 organizes related targets and covariates through group attention~\citep{arango2025chronosx,ansari2025chronos2}. Collectively, these studies establish that population-level parameter sharing and explicit cross-series interaction are complementary. The remaining retail-specific design concerns how product groups, regional coverage, covariate roles, and forecast-time availability determine admissible sharing. \model{} addresses that organization in a task-trained setting, without equating it with general-purpose zero-shot transfer.

\subsection{Language-conditioned Forecasting}
Language-based forecasting spans numerical serialization, pretrained-backbone adaptation, and semantic conditioning. LLMTime treats numerical strings as language-model continuations~\citep{gruver2023llmtime}; GPT4TS (One Fits All) and Time-LLM adapt pretrained language computation to numerical forecasting~\citep{zhou2023onefitsall,jin2024timellm}. UniTime uses domain instructions for cross-domain learning~\citep{liu2024unitime}, while TimeCMA aligns time-series and prompt representations~\citep{liu2025timecma}. These mechanisms transfer representations, but the content and incremental value of the text remain separate questions.

Multimodal resources and models make this distinction explicit. Time-MMD pairs numerical and textual data, Context is Key evaluates forecasts requiring textual knowledge or constraints, and CAPTime connects exogenous context to probabilistic forecasting~\citep{liu2024timemmd,williams2025context,yao2025captime}. Together with controlled analyses of LLM-based forecasters~\citep{tan2024useful}, this literature motivates distinguishing additional business information from architectural capacity or repeated historical inputs.

Our emphasis is the transfer of flexibly expressed plans and experience into target- and date-specific numerical prediction. Language hidden representations are aligned with the target's memory and condition future queries; sales supervision learns the response. This positions the interface as learned business-context conditioning. Matching numerical evidence, text content, and forecast-time visibility in evaluation is essential to interpreting that contribution.

\subsection{Regional Collaboration and Hierarchical Forecasting}
Hierarchical forecasting imposes relationships among outputs. Optimal-combination and minimum-trace methods reconcile forecasts under aggregation constraints~\citep{hyndman2011hierarchical,wickramasuriya2019mint}, while end-to-end probabilistic models incorporate coherence into learning~\citep{rangapuram2021hierarchical}. Their common objective is consistency across a defined hierarchy.

Regional representation sharing determines which histories provide useful evidence, with group attention supplying a general interaction mechanism~\citep{ansari2025chronos2}. Related representation-learning studies explore information sharing under sparse observations~\citep{han2023crossdomain,han2023dominantsets,han2024transport}. Retail transfer learning also uses category-level sales to regularize product-level demand prediction~\citep{lei2025pooling}. The distinction is consequential: shared parameters transfer population patterns, regional messages transfer target-specific evidence, and reconciliation constrains output sums. \model{} uses coverage-aware, directed exchange for the second role. Incomplete DC observations and service footprints do not themselves establish an additive label hierarchy, so national/DC collaboration preserves separate regional outputs without imposing a national-equals-DC-sum constraint.

\subsection{Conditional Specialization and Planning Quantities}
Sparse conditional computation separates model capacity from the computation activated for each input~\citep{shazeer2017moe,fedus2022switch}. Moirai-MoE applies this principle to time-series specialization~\citep{liu2024moirai}. The shared/routed decoder in \model{} uses these ideas to adapt transformations to product, seasonal, activity, and regional context; experts have no predefined one-to-one business assignment.

\Needspace{3\baselineskip}
Quantized forecasting and proper scoring rules provide foundations for probabilistic outputs~\citep{ansari2024chronos,gneiting2007scoring}, while temporal hierarchies connect prediction across aggregation levels~\citep{athanasopoulos2017temporal}. Context-masking approaches address incomplete histories and parallel forecasting~\citep{auer2025tirex,khwaja2026toto2}. These objectives are related but distinct: distributional learning represents uncertainty, temporal aggregation concerns planning quantities, and masking concerns available evidence. Related decision models learn replenishment policies directly~\citep{qi2023inventory} and integrate assortment planning with inventory allocation~\citep{shen2025fulfillment}. ORPR pretrains replenishment policies on optimization-derived reference decisions and refines them through reinforcement learning~\citep{zhao2026orpr}. Our daily/window supervision constrains sums from the same daily forecast trajectory while retaining daily distribution learning.

These strands motivate the computation developed next: select admissible numerical evidence, ground business language in that evidence, and decode target-specific daily forecasts and planning quantities.

\section{TimeHF2.0}
\label{sec:method}
Building on Section~\ref{sec:related}, \model{} connects numerical encoding, temporal and directed cross-stream attention, text alignment, and parallel decoding. The computation gives business knowledge a target- and date-specific path into daily forecasts, with planning-window supervision connecting those forecasts to cumulative quantities.

\begin{figure}[t]
\centering
\includegraphics[width=0.98\linewidth]{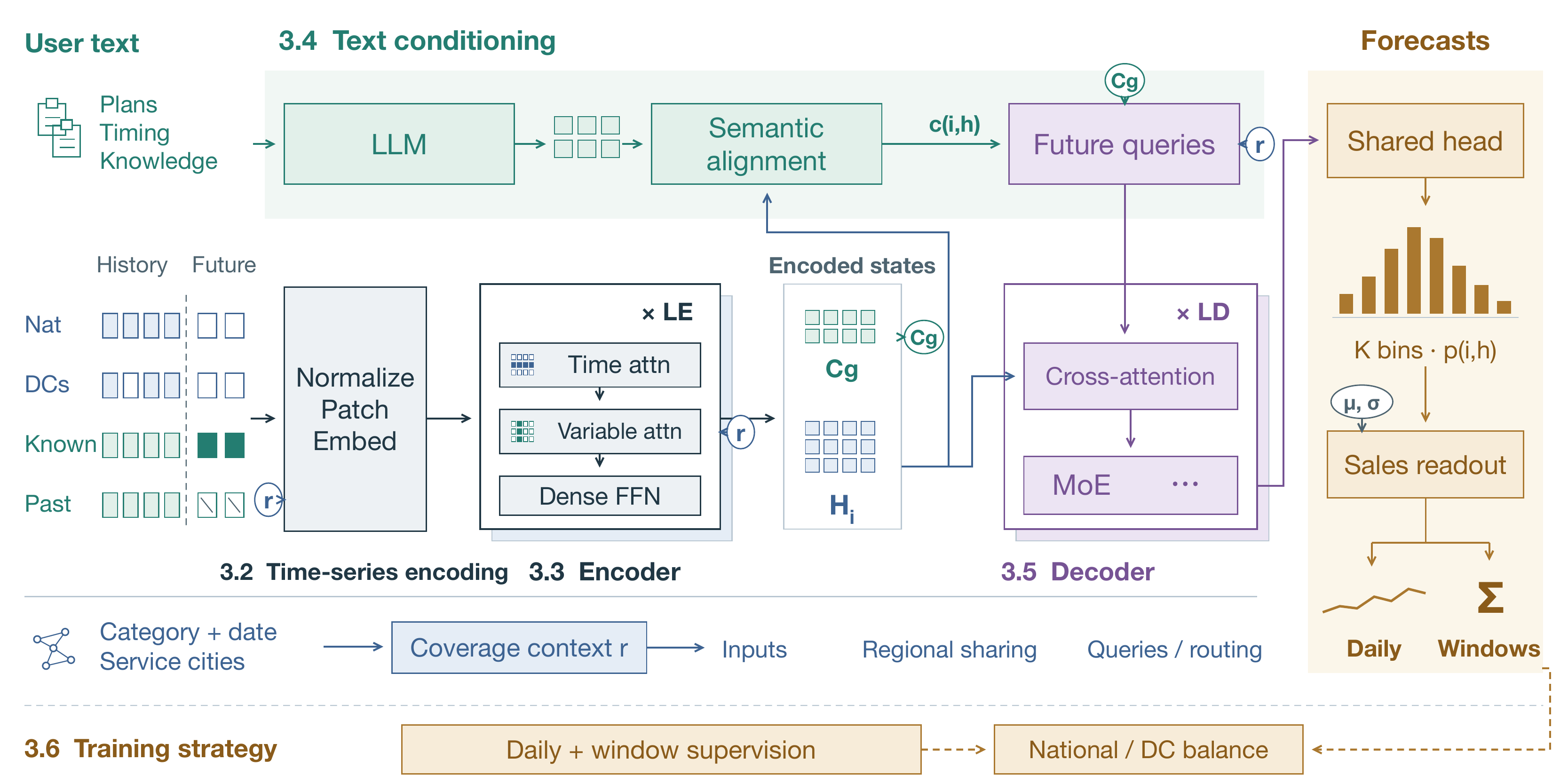}
\caption{\small \textbf{TimeHF~2.0 framework and section correspondence.} Section~\ref{sec:encoding} prepares time-series inputs and coverage context; Section~\ref{sec:encoder} applies temporal attention, directed variable attention, and FFN to form $H_i$ and $\mathsf C_g$. Section~\ref{sec:multimodal} transfers language-model hidden representations of users' plans and business knowledge through semantic alignment with numerical memory into context $c_{i,h}$. Section~\ref{sec:decoder} applies that context in future queries, reads the original $H_i$, and routes the text-conditioned decoding state through shared/routed expert processing and a shared head for daily/window forecasts. The dashed supervision path corresponds to Section~\ref{sec:training}; $r$ and $\mathsf C_g$ ports denote regional and covariate context. Token grids and expert/probability symbols are schematic.}
\label{fig:architecture}
\end{figure}

\subsection{Problem Definition and Overview}
\label{sec:task}
At forecast origin $T$, a product group $g$ contains national sales, the applicable distribution-center (DC) sales targets, and relevant covariates. For each target $i$, the model uses history through $T$ and origin-visible conditions to predict daily sales over $T+1,\ldots,T+H$. Conditions include available future schedules and user-supplied descriptions of plans, activity timing, and business experience, including qualitative expectations about local conditions. The horizon covers the future period relevant to inventory allocation and supply-chain planning. Future sales are used only as labels. Parameters are shared across product groups, while explicit numerical-stream exchanges remain within a group.

Figure~\ref{fig:architecture} follows the same organization as this section. Time-series encoding constructs aligned input states; the encoder produces each target's numerical memory $H_i$ and covariate states $\mathsf C_g$. Text conditioning aligns language-model hidden representations with $H_i$ to form context $c_{i,h}$ for the relevant target and future date. The decoder uses this context to condition future queries, reads the original $H_i$, and applies shared and routed experts before a shared prediction head. Daily point forecasts also determine the cumulative quantities used in procurement and replenishment windows. The lower dashed path shows the training supervision rather than an additional forward prediction module.

Common parameters learn population patterns across products, whereas within-group attention retrieves evidence for a particular product and origin. A DC can use the same product's national trajectory while retaining its own sales scale, observations, and coverage. This connects C1 and C2: the model must learn which reference is useful and how its information applies to the receiving target.

\Needspace{16\baselineskip}
\subsection{Time-series Encoding}
\label{sec:encoding}
\label{sec:availability}
This stage converts numerical observations and schedules into the stream states that the encoder can process. Sales targets, future-available covariates, and past-only covariates retain their roles and scope. Each numerical stream uses its own retained historical statistics; sales use a normalized asinh transform to compare magnitudes across products and regions. Historical values and observation masks form patches, followed by projection and type, role, and time encoding. Future dates use aligned daily slots: target sales have learned placeholders, known covariates contain their available values, and unavailable inputs are masked. Zero sales are valid observations, distinct from missing values and padding.

Regional context $r_{i,t}$ represents the market served by a target. Category- and date-conditioned attention pools its service-city embeddings, together with coverage identity, size, and missingness. The national target has a national-level representation. This context enters target input states and remains available for regional sharing, future queries, and routing. A shared covariate is encoded once per applicable group; individual targets can still learn different responses. Thus this stage preserves the meaning and availability of each input without yet exchanging hidden states between streams.

Input roles specify what can be known, separately from whether a value happens to be present in a stored record. A common campaign schedule can apply to all participating products across the national and DC targets, while a weather forecast describes the conditions expected for a target's service area. The shared campaign need not produce identical responses at all targets. Coverage uses the service footprint known at $T$; pooled city weights describe contextual relevance rather than measured sales shares. Keeping these roles explicit allows the encoder to combine shared conditions with target-specific histories.

\Needspace{14\baselineskip}
\subsection{Encoder: Temporal and Directed Variable Attention}
\label{sec:encoder}
The encoder transforms the prepared streams into numerical evidence for forecasting. Each layer has three steps: time-axis attention, directed variable-axis attention, and a dense feed-forward network (FFN). Suppressing position indices where unambiguous,
\begin{equation}
\begin{aligned}
U_s^{(\ell)}&=X_s^{(\ell)}+
\operatorname{TimeAttn}_{\ell}\!\left(\LN(X_s^{(\ell)})\right),\\
V_i^{(\ell)}&=U_i^{(\ell)}+
\mathcal M_Y^{(\ell)}(U^{(\ell)};r)_i+
\mathcal M_C^{(\ell)}(U^{(\ell)};\mathcal A_T)_i,\\
X_i^{(\ell+1)}&=V_i^{(\ell)}+
\operatorname{FFN}_{\ell}\!\left(\LN(V_i^{(\ell)})\right).
\end{aligned}
\label{eq:encoding}
\end{equation}
Here $X^{(0)}$ is the time-series encoding and $\mathcal A_T$ describes permitted roles, scope, and origin-time availability. The first line models temporal dependencies separately within every sales or covariate stream, using admissible history and known future conditions. It produces $U$, the common pre-update states used by the variable-axis step.

At each aligned position, a sales target reads two families of sources. The regional branch $\mathcal M_Y$ reads valid national and peer-DC sales, transferring useful patterns while retaining the receiving target's own state. Source reliability describes the reference history, pairwise compatibility reflects its relevance to the receiving region, and a receiver gate controls how much information enters. Coverage context helps distinguish a relevant reference from a well-observed but different market. Sparse sales alone do not imply unreliable evidence.

These controls answer different questions. Reliability concerns whether the source observations support a useful read; compatibility concerns whether the source is relevant to this receiving market; the receiver gate concerns how strongly to use the resulting message. For example, a short DC history may provide little evidence about the timing of a seasonal rise, while national and peer histories contain a clearer transition. Regional attention can supply that temporal reference, with the DC's own state and scale preserving its local level. Conversely, a well-observed peer need not be a useful reference when its coverage and dynamics differ.

The covariate branch $\mathcal M_C$ reads conditions applicable to the target, such as a shared activity schedule or a regional weather-forecast stream available at the origin. Both branches read the same $U$ in parallel, with shared variable-attention projections but separate source masks and normalizations. Each first forms a weighted read over its permitted sources, followed by its gated residual contribution. Consequently, regional references do not mechanically dilute covariate attention when the number of DCs changes. Covariate readers can read permitted condition streams but cannot read target hidden states; their regional-target branch is absent. This preserves directed sales--covariate modeling across layers.

The FFN then transforms each stream-position state independently, without another exchange between regions or variables. Repeating the three-step layer forms $H_i$, which combines the target's temporal history, selected national/DC evidence, and relevant business conditions. The gating and null-source conventions are specified in Appendix~\ref{app:attention}.

\subsection{Text Conditioning from Flexible Plans and Business Knowledge}
\label{sec:multimodal}
Users can describe future plans and their intended rhythm in natural language: which products participate, when a promotional lead-in starts, when promotion effort is concentrated, or how an activity changes across phases. They can also supply qualitative business experience about consumption occasions and likely demand responses. These descriptions supplement structured schedules with relations and meaning that need not be expressed as a fixed inventory of numerical features.

A user can describe the forecasting background and a campaign update:
\begin{quote}
\small\textbf{Background.} Forecast national sales for a campaign-sensitive product over the upcoming planning horizon.\par
\textbf{User update.} Schedule promotional lead-in for $T+4$--$T+8$, main activity for $T+9$--$T+15$, and follow-up for $T+16$--$T+21$. Expect demand to build during the lead-in and ease after the main activity.
\end{quote}
The background identifies the product and target, while the update specifies the campaign phases and expected response. This knowledge provides context that a numerical history alone may not reveal. Users can add or revise it directly and obtain conditioned forecasts without hand-coding a demand coefficient or retraining a separate model for each update. The numerical response is learned from forecast supervision.

The input interface separates a concise task background from the new business update. Background identifies the product type, target, forecast period, and retained numerical evidence; the update states the plan or expectation and its applicable dates. These fields organize scope and timing without requiring a fixed sentence template. Historical sales included in the text follow the same origin and masks as the numerical branch. Thus a newly supplied campaign description adds business context to an existing forecast, while repeating history in prose does not create additional observations.

A large language model (LLM) produces hidden representations of the description together with the relevant product and regional background. A trainable semantic alignment module transfers these representations into the forecasting space and connects them with target memory $H_i$. For description unit $j$, let $Z_{i,j}$ denote its projected language hidden states. The connection is summarized as
\begin{equation}
\begin{aligned}
\widetilde Z_{i,j}&=\operatorname{Align}(Z_{i,j},H_i),\\
c_{i,h}&=\operatorname{Aggregate}_{j\in\mathcal J_{i,h}}\!\left(\widetilde Z_{i,j}\right).
\end{aligned}
\label{eq:text_memory}
\end{equation}
Alignment uses cross-attention: projected text representations query the target's numerical memory, retaining the semantic input while incorporating relevant sales evidence. This gives a statement about an occasion or activity a connection to the product's observed dynamics. The set $\mathcal J_{i,h}$ contains origin-visible base-context and user-description units applicable to target $i$ and date $T+h$. Relative dates follow the forecast calendar, and applicability uses the declared response period. Aggregate pools the valid aligned text states into the forecasting-dimensional context vector $c_{i,h}$; an empty applicable set contributes zero text context.

The module therefore links three kinds of information: what business users expect or intend, which target and future period they mean, and what the numerical history supports. A revised activity phase, an expected weather change, or category experience can enter through the same interface and influence both the future sales trajectory and its planning-window totals. Forecast supervision learns how this aligned context should be used.

\subsection{Decoder: Text-conditioned Queries and Mixture of Experts}
\label{sec:decoder}
Text-to-memory alignment grounds a description in the target's dynamics; the decoder then uses that context to retrieve evidence for each future date. These complementary reads allow a shared business note to interact with different regional histories. Each base query $q^{\mathrm{base}}_{i,h}$ combines target-history summary, forecast lead time, category, regional context, and applicable future covariates. A learned scalar gate $\beta_{i,h}$ depends on this query and $c_{i,h}$, so semantic context can have different influence on different future dates. Query construction and each decoder layer are
\begin{equation}
\begin{aligned}
q^{(0)}_{i,h}&=q^{\mathrm{base}}_{i,h}+\beta_{i,h}c_{i,h},\\
u^{(\ell)}_{i,h}&=q^{(\ell)}_{i,h}+
\operatorname{CrossAttn}_{\ell}\!\left(\LN(q^{(\ell)}_{i,h}),H_i\right),\\
q^{(\ell+1)}_{i,h}&=u^{(\ell)}_{i,h}+
\eta_\ell\operatorname{MoE}_{\ell}\!\left(\LN(u^{(\ell)}_{i,h});\xi_{i,h}\right).
\end{aligned}
\label{eq:decoder}
\end{equation}
Here $\beta_{i,h}=\operatorname{sigmoid}(w_\beta^\top[q^{\mathrm{base}}_{i,h};c_{i,h}]+b_\beta)$. The original $H_i$ remains the decoder's K/V, with projections separate from cross-modal attention. Thus a user's planned phases or business expectation affect the future query, its numerical-memory read, and the subsequent expert transformation. All future dates are processed in parallel, without future-query self-attention or feedback of predicted sales.

Each layer combines shared processing with a sparsely activated set of routed experts:
\begin{equation}
\operatorname{MoE}(x;\xi)=E_{\mathrm{sh}}(x)+
\sum_{e\in\mathcal E_{\mathrm{act}}(x,\xi)}\widetilde\pi_e E_e(x).
\label{eq:moe}
\end{equation}
Shared processing provides a common transformation; selected experts add contextual capacity~\citep{liu2024moirai}. Router probabilities $\pi$ determine $\mathcal E_{\mathrm{act}}$, and $\widetilde\pi$ renormalizes the selected probabilities. The router reads decoder state $x$ together with $\xi_{i,h}$, which collects category, calendar, regional, and applicable structured-covariate context. Semantic information is already present in $x$ through the text-conditioned query and numerical-memory read. Thus plans and experience influence expert selection through the conditioned decoding state. Experts learn context-dependent hidden-state transformations, retain a shared output head, and have no fixed one-to-one assignment to categories or events. The residual coefficient $\eta_\ell$ controls their contribution in Equation~\ref{eq:decoder}.

The shared head uses a categorical forecasting formulation~\citep{ansari2024chronos}, predicting $p_{i,h,k}$ over $K$ ordered bins. Our representation applies these bins to normalized, asinh-transformed sales. The nonuniform bins preserve fine detail in routine variation and use wider tails for extreme sales levels. This is useful when an occasional bulk order produces an isolated peak without a matching origin-visible covariate: observations within a tail bin do not require distinct exact-amplitude labels. A promotion-driven rise remains learnable through the covariate and text paths. All valid sales observations, including genuine high-demand periods, continue to supervise learning. The point-regression penalty and window supervision retain numerical and accumulated-quantity constraints; Section~\ref{sec:robust_eval} evaluates this combination. With finite bin centers $\nu_k$ and target history statistics $\mu_i,\sigma_i$,

\begin{equation}
\begin{aligned}
v_{i,k}&=\max(0,\mu_i+\sigma_i\sinh(\nu_k)),\\
\widehat y_{i,h}&=\sum_{k=1}^{K}p_{i,h,k}v_{i,k},\qquad
\widehat S_{i,a,w}=\sum_{h=a}^{a+w-1}\widehat y_{i,h}.
\end{aligned}
\label{eq:decode}
\end{equation}
Representatives are restored to sales units before probability averaging, and planning totals use the same daily point forecasts. National and DC targets retain separate outputs; no hard national-equals-DC-sum constraint is imposed. Appendix~\ref{app:bins} defines the bins and finite-range implications.

This readout connects C3 and C4. Expert transformations adapt the hidden representation to the current forecasting conditions, while the shared head places their outputs on the same ordered sales scale. A revised plan can change the timing, level, or shape of daily predictions, and its effect on a planning interval follows directly from their sum. The model therefore learns a single trajectory whose daily precision and accumulated quantity can be assessed together.

\subsection{Training Strategy}
\label{sec:training}
\label{sec:objectives}
Forecast supervision jointly learns the numerical forecasting computation and the semantic connection. The alignment and query-conditioning paths learn how descriptions of experience, plans, and changes relate to future sales when combined with historical evidence. Future sales supervise both daily distributions and planning-window quantities:
\begin{equation}
\begin{aligned}
\mathcal L_i&=\overline{\mathcal L}_{\mathrm{CE},i}
+\lambda_{\mathrm{ord}}\overline{\mathcal L}_{\mathrm{RPS},i}
+\lambda_{\mathrm{reg}}\mathcal L_{\mathrm{reg},i}
+\lambda_{\mathrm{win}}\mathcal L_{\mathrm{win},i},\\
\mathcal L_{\mathrm{win},i}&=\operatorname{Mean}_{(a,w)\in\mathcal W_i}
\operatorname{Huber}_{\delta}\!\left[\log(1+\widehat S_{i,a,w})-\log(1+S_{i,a,w})\right].
\end{aligned}
\label{eq:learning}
\end{equation}
Cross-entropy (CE) learns true-bin probability, while the ranked probability score (RPS) accounts for ordered sales levels~\citep{gneiting2007scoring}; together they form ordered-bin supervision. The point-regression term $\mathcal L_{\mathrm{reg},i}$ penalizes discrepancies between decoded daily forecasts and valid daily sales labels, preserving numerical calibration within the coarse representation. Its penalty and weight are training configuration choices. Window supervision checks accumulated quantities from the same daily trajectory. Daily terms average valid labels; $\mathcal W_i$ contains fully labeled intervals within the $H$-day horizon. The target sum $S_{i,a,w}$ uses the same interval. Within each product group, the mean valid-DC loss and national loss receive configurable weights, followed by averaging across groups. The full objective also includes an expert-balancing term (Appendix~\ref{app:learning}).

The loss terms constrain complementary aspects of that trajectory. Daily CE and RPS distinguish ordered sales levels, the point-regression penalty constrains decoded quantities, and window loss penalizes persistent bias across dates. Regression and cumulative penalties must be balanced against the bin terms so that isolated amplitudes do not dominate the combined objective. A trajectory with an approximately correct total can still have poor daily timing, while small daily deviations can accumulate into a materially incorrect replenishment quantity. Neither objective replaces the other. Averaging valid DC losses within a group also prevents a product from receiving greater supervision weight simply because more DC targets are present.

Training combines complete context with contiguous history masking~\citep{auer2025tirex}: independent masks retain regional references when a target interval is hidden, while synchronized masks simulate missing group context. Masks precede numerical statistics, patches, and textual sales lists or summaries. All plans and descriptions use versions available at $T$; user experience is an input belief, not a future observation. Training also covers disabled-language samples. At a fixed origin and with fixed numerical inputs, revising a note leaves the scalers and encoder memory unchanged. It changes text alignment, future queries, and downstream decoding; the same learned parameters produce revised daily forecasts and window totals. Changes to the origin or numerical conditions rebuild the corresponding numerical record. Section~\ref{sec:experiments} describes the industrial setting and evaluates the corresponding national, DC, and business-knowledge capabilities.

\section{Industrial Evaluation}
\label{sec:experiments}
The evaluation connects regional information, business conditions, and the learning objectives in Section~\ref{sec:method} to daily forecasting and cumulative inventory-planning quantities. We examine national and DC results, component contributions, robustness to retail volatility, and performance across product types, followed by business-knowledge and regional collaboration cases.

\subsection{Industrial Setting and Evaluation}
\label{sec:system}
\label{sec:eval_protocol}
The online experiments evaluate 200,000 SKUs from JD's retail catalogue during January--September 2026. Training draws on a corpus of 912 billion data points sampled from retail records spanning \mbox{2023--2025}. National forecasting evaluates these products at the national level, while DC forecasting evaluates their applicable product--DC targets. Candidate models and the original production deep learning baseline $B$ use matched product--region--origin records and evaluation windows. TimeHF~1.0 provides an additional model comparison.

The 380M numerical forecasting backbone is paired with a language model from a 0.8B, 9B, and 27B family post-trained by our team to strengthen key capabilities for large-scale retail, including forecasting. The evaluated model uses 0.8B. A forthcoming technical report will detail this family.

At each forecast origin $T$, the model predicts daily sales over a future horizon $H$ and accumulated sales over the planning intervals used for inventory allocation, replenishment, and supply-chain management. Daily accuracy measures agreement at individual dates; planning-window accuracy measures the quantity available to support these operational decisions.

Inputs use origin-visible sales histories, structured conditions, and applicable text descriptions. Campaign schedules use the version known at $T$, weather inputs use forecasts already issued, and historical page views (PV) describe past customer activity. Text specifies product/region scope, event dates, and response intervals. Zero sales remain valid observations, with missingness and padding handled separately. Observed sales can be constrained by stock availability. Isolated large orders remain in the evaluation record, while robust learning controls their influence during fitting.

For evaluation records $i\in\mathcal G$, let $S_i$ and $\widehat S_i^M$ denote actual and predicted quantities for model $M$. Daily evaluation uses daily quantities; planning-window evaluation sums each target's daily quantities over the relevant interval. Accuracy and relative accuracy improvement are
\begin{equation}
\begin{aligned}
A(M;\mathcal G)&=1-\frac{\sum_{i\in\mathcal G}|\widehat S_i^M-S_i|}{\sum_{i\in\mathcal G}S_i},\\
\operatorname{RAI}(M;B,\mathcal G)&=100\,\frac{A(M;\mathcal G)-A(B;\mathcal G)}{A(B;\mathcal G)}\%.
\end{aligned}
\label{eq:online_metrics}
\end{equation}
The numerator of $A$ sums target-level absolute deviations. National and DC targets are aggregated separately, with positive actual totals and baseline accuracy. The result tables report relative accuracy improvement (\%) over $B$.

\subsection{National Forecasting Results}
\label{sec:overall}
Table~\ref{tab:overall} compares national daily and planning-window accuracy. The two measures assess the timing and level of daily sales and the accumulated quantity used in procurement and supply planning.

\begin{table}[H]\centering\small
\caption{National online forecasting results. Entries are relative accuracy improvement (\%).}
\label{tab:overall}
\setlength{\tabcolsep}{10pt}\renewcommand{\arraystretch}{1.15}
\begin{tabularx}{.90\linewidth}{@{}Xrr@{}}\toprule
& \multicolumn{2}{c}{Relative accuracy improvement (\%)}\\
\cmidrule(l){2-3}
Model & Daily & Planning window\\\midrule
Original DL baseline & --- & ---\\
TimeHF~1.0 & \NationalOneDailyGain & \NationalOneWindowGain\\
\rowcolor{SCOAIPale}\textbf{TimeHF~2.0} & \textbf{\NationalTwoDailyGain} & \textbf{\NationalTwoWindowGain}\\\bottomrule
\end{tabularx}
\end{table}

\model{} achieves the strongest results at both granularities: \NationalTwoDailyGain\% in daily accuracy and \NationalTwoWindowGain\% in planning-window accuracy, compared with \NationalOneDailyGain\% and \NationalOneWindowGain\% for TimeHF~1.0. The improvement in daily prediction carries through to the cumulative planning quantity, indicating that the combined regional and business-conditioned representation benefits both short-term sales tracking and inventory decisions.

\subsection{DC Daily and Planning-window Results and Ablations}
\label{sec:ablations}
Table~\ref{tab:ablations} compares the complete model with TimeHF~1.0 and four ablated configurations. The covariate ablation excludes structured covariate inputs and their conditioning paths, while the cumulative ablation omits the window-loss term. In the regional ablation, national and DC targets rely on their own sales histories without cross-region exchange. The multimodal ablation has no text-conditioning module or associated decoder injection.

\begin{table}[H]\centering\small
\caption{DC online results and component ablations. Entries are relative accuracy improvement (\%).}
\label{tab:ablations}
\setlength{\tabcolsep}{9pt}\renewcommand{\arraystretch}{1.13}
\begin{tabularx}{\linewidth}{@{}Xrr@{}}\toprule
& \multicolumn{2}{c}{Relative accuracy improvement (\%)}\\
\cmidrule(l){2-3}
Model / variant & Daily & Planning window\\\midrule
Original DL baseline & --- & ---\\
TimeHF~1.0 & \DCOneDailyGain & \DCOneWindowGain\\
\rowcolor{SCOAIPale}\textbf{TimeHF~2.0} & \textbf{\DCFullDailyGain} & \textbf{\DCFullWindowGain}\\
\addlinespace[3pt]
\quad w/o covariate information & \DCNoCovariateDailyGain & \DCNoCovariateWindowGain\\
\quad w/o cumulative supervision & \DCNoCumulativeDailyGain & \DCNoCumulativeWindowGain\\
\quad w/o regional information & \DCNoRegionDailyGain & \DCNoRegionWindowGain\\
\quad w/o multimodal information & \DCNoMultimodalDailyGain & \DCNoMultimodalWindowGain\\\bottomrule
\end{tabularx}
\end{table}

The complete model achieves \DCFullDailyGain\% daily and \DCFullWindowGain\% planning-window improvement, exceeding TimeHF~1.0's \DCOneDailyGain\% and \DCOneWindowGain\%. All four removals weaken both measures. Removing covariates gives the lowest ablation results, \DCNoCovariateDailyGain\% daily and \DCNoCovariateWindowGain\% over the planning window, highlighting the importance of applicable business conditions. Removing cumulative supervision gives \DCNoCumulativeDailyGain\% and \DCNoCumulativeWindowGain\%, showing that direct supervision of planning quantities also improves the underlying daily trajectory.

Without regional information, the daily and window improvements fall to \DCNoRegionDailyGain\% and \DCNoRegionWindowGain\%. This supports transferring useful national and peer-DC patterns to local targets. Removing multimodal information yields \DCNoMultimodalDailyGain\% and \DCNoMultimodalWindowGain\%; flexible business descriptions therefore provide an additional benefit alongside structured covariates and numerical histories. The consistent ordering of the ablations in both columns supports complementary contributions from information integration and planning-oriented learning.

\subsection{Ordered-bin Learning under Retail Volatility}
\label{sec:robust_eval}
Retail sales contain both learnable demand changes and isolated transaction extremes. A scheduled promotion provides an available business condition for learning a sustained lift. By contrast, one-off large orders from individual retail customers, repeated bulk purchases, and uncontrolled order concentration in private customer channels can produce sharp peaks without matching campaign, PV, or weather signals.

\begin{figure}[t]\centering
\includegraphics[width=\linewidth]{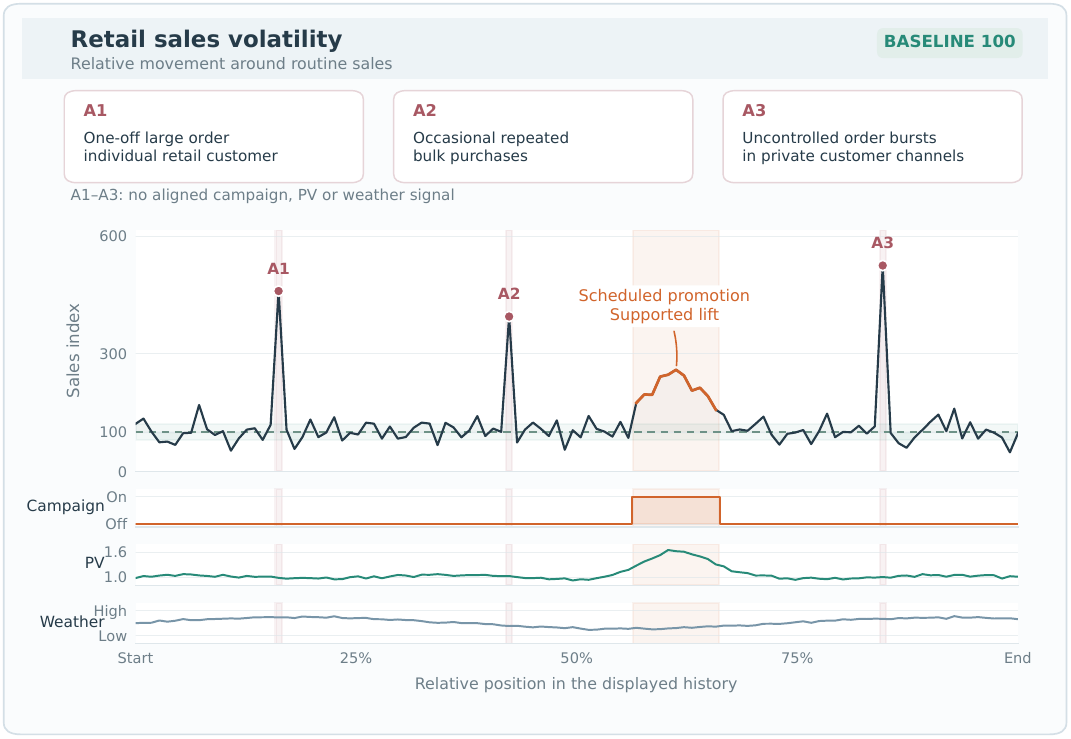}
\caption{\textbf{Online retail sales fluctuations around a routine baseline of 100.} Sales are indexed as $100\times y_t/b$, where $b$ is the routine baseline. A1 denotes a one-off large order from an individual retail customer; A2, occasional repeated bulk purchases; and A3, uncontrolled order concentration in private customer channels. These isolated peaks have no matching campaign, PV, or weather signal. In contrast, the shaded promotion has an available campaign signal and elevated PV. Learning should preserve this supported lift while limiting the influence of isolated extreme amplitudes.}
\label{fig:volatility_case}
\end{figure}

\Needspace{4\baselineskip}
Figure~\ref{fig:volatility_case} contrasts these isolated peaks with a promotion-supported rise in online retail sales. Learning should capture reproducible demand responses while limiting the influence of unpredictable peak amplitudes.

\Needspace{10\baselineskip}
We compare continuous regression heads trained with mean squared error (MSE), mean absolute error (MAE), or Huber loss against ordered-bin prediction. Continuous heads predict daily sales directly; the ordered-bin head follows categorical forecasting~\citep{ansari2024chronos}. Bin-only supervision combines CE and RPS; the full objective adds regression and cumulative supervision (Equation~\ref{eq:learning}). Table~\ref{tab:losses} compares these output representations and supervision schemes on national and DC planning-window accuracy. Table~\ref{tab:ablations} separately removes cumulative supervision from the complete model.

\begin{table}[H]\centering\small
\caption{Output-head and training-objective comparison. Entries are planning-window relative accuracy improvement (\%).}
\label{tab:losses}
\setlength{\tabcolsep}{10pt}\renewcommand{\arraystretch}{1.13}
\begin{tabularx}{\linewidth}{@{}Xrr@{}}\toprule
& \multicolumn{2}{c}{Window relative accuracy improvement (\%)}\\
\cmidrule(l){2-3}
Output / supervision & National & DC\\\midrule
Continuous regression: MSE & \LossMSENational & \LossMSEDC\\
Continuous regression: MAE & \LossMAENational & \LossMAEDC\\
Continuous regression: Huber & \LossHuberNational & \LossHuberDC\\
Ordered-bin head: CE + RPS & \LossBinsNational & \LossBinsDC\\
\rowcolor{SCOAIPale}\textbf{Ordered-bin head: full objective} & \textbf{\NationalTwoWindowGain} & \textbf{\DCFullWindowGain}\\\bottomrule
\end{tabularx}
\end{table}

\Needspace{5\baselineskip}
Table~\ref{tab:losses} shows stronger continuous-regression results with MAE and Huber than with MSE. MSE gives \LossMSENational\% national and \LossMSEDC\% DC improvement, MAE gives \LossMAENational\% and \LossMAEDC\%, and Huber gives \LossHuberNational\% and \LossHuberDC\%. MSE gradients grow with residual magnitude, whereas MAE and Huber limit the pull of isolated extremes~\citep{huber1964robust}, helping the model retain representative sales behavior under volatile observations.

The ordered-bin head trained with CE and RPS further improves the results to \LossBinsNational\% nationally and \LossBinsDC\% at DC level. Fine central bins preserve routine detail, while wider tails represent extreme amplitudes coarsely. This allows learning of a high-sales regime without reproducing every isolated peak exactly; covariate-supported promotion responses remain learnable through the conditioning paths.

\Needspace{4\baselineskip}
The full objective achieves the best results, \NationalTwoWindowGain\% nationally and \DCFullWindowGain\% at DC level. The regression penalty calibrates decoded quantities, while cumulative supervision aligns the trajectory with inventory-planning totals. Their combination balances robust amplitude learning with useful numerical detail and accumulated sales.

\subsection{Performance across Product Types}
\label{sec:product_types}
Retail products differ in sales volume, seasonality, intermittency, and historical coverage. Table~\ref{tab:product_types} compares planning-window performance across bestsellers, seasonal products, long-tail products, and new products at both national and DC levels.

\begin{table}[H]\centering\small
\caption{Planning-window relative accuracy improvement (\%) across product types. Baseline columns refer to the original production deep learning model.}
\label{tab:product_types}
\setlength{\tabcolsep}{5pt}\renewcommand{\arraystretch}{1.18}
\begin{tabularx}{\linewidth}{@{}Xrrrrrr@{}}\toprule
& \multicolumn{3}{c}{National} & \multicolumn{3}{c}{DC}\\
\cmidrule(lr){2-4}\cmidrule(l){5-7}
Product type & Baseline & TimeHF~1.0 & TimeHF~2.0 & Baseline & TimeHF~1.0 & TimeHF~2.0\\\midrule
Bestsellers & --- & \BestOneNational & \textbf{\BestTwoNational} & --- & \BestOneDC & \textbf{\BestTwoDC}\\
Seasonal & --- & \SeasonOneNational & \textbf{\SeasonTwoNational} & --- & \SeasonOneDC & \textbf{\SeasonTwoDC}\\
Long-tail & --- & \TailOneNational & \textbf{\TailTwoNational} & --- & \TailOneDC & \textbf{\TailTwoDC}\\
New products & --- & \NewOneNational & \textbf{\NewTwoNational} & --- & \NewOneDC & \textbf{\NewTwoDC}\\\bottomrule
\end{tabularx}
\end{table}

\model{} outperforms TimeHF~1.0 in every product group at both target levels. Bestsellers reach \BestTwoNational\% national and \BestTwoDC\% DC improvement, indicating benefits even when sales observations are relatively rich. The strongest DC improvements occur for seasonal and long-tail products: \SeasonTwoDC\% and \TailTwoDC\%, compared with \SeasonOneDC\% and \TailOneDC\% for TimeHF~1.0. Regional evidence is particularly useful when a local seasonal transition is weakly observed or sales are intermittent, because national and peer histories supply additional timing and regime information.

New products improve from \NewOneNational\% to \NewTwoNational\% nationally and from \NewOneDC\% to \NewTwoDC\% at DC level. Shared representations and contextual information help compensate for short local histories. Across all four product groups, the reported DC gains exceed the national gains, with the largest separation for seasonal and long-tail products, consistent with the value of regional information for local forecasting.

The architecture combines broadly reusable transformations with context-dependent specialization. Shared experts learn common forecasting behavior, while routed experts adapt processing to product, season, activity, and region. National--DC collaboration supplies complementary numerical evidence to these transformations. This combination supports differentiated predictions across product regimes while preserving each receiving target's own sales scale and temporal behavior.

\subsection{Business-knowledge Conditioning and Case Studies}
\label{sec:knowledge_eval}
\label{sec:knowledge_cases}
The multimodal branch improves both temporal detail and accumulated planning quantity. In Table~\ref{tab:ablations}, the complete model obtains \DCFullDailyGain\% daily and \DCFullWindowGain\% window improvement; removing multimodal information lowers these results to \DCNoMultimodalDailyGain\% and \DCNoMultimodalWindowGain\%. The improvement at both granularities shows that business meaning benefits the timing and accumulated quantity of forecasts.

\newcommand{\KnowledgeCasesIntro}{%
Figures~\ref{fig:case_school}--\ref{fig:case_fathers_day} show how adding or revising campaign and occasion knowledge changes forecast timing and level. Paired conditions share their numerical inputs and product background. Shading marks the activity or lead-in period; the sales index uses a retained historical mean of 100, and $T+H$ marks the forecast-window endpoint.
\par\medskip
}
\newcommand{\KnowledgeOutcomeLabel}{Observed sales}
\newcommand{\KnowledgeCasePlot}[8][4.2cm]{%
\ReadCaseTable{data/#2_plot.csv}%
\pgfmathsetmacro{\NewWindowLeft}{#4-.5}%
\pgfmathsetmacro{\NewWindowRight}{#5+.5}%
\pgfmathsetmacro{\OldWindowLeft}{#7-.5}%
\pgfmathsetmacro{\OldWindowRight}{#8+.5}%
\begin{tikzpicture}
\begin{axis}[case axis,width=.96\linewidth,height=#1,
  xmin=-20,xmax=28,ymax=#3,ylabel={Sales index},
  xtick={-20,0,14,28},xticklabels={$T-L$,$T$,$T+H/2$,$T+H$},
  legend style={font=\scriptsize,draw=none,fill=none,
    at={(.5,-.28)},anchor=north,legend columns=3,column sep=7pt}]
\ifnum#7>0\relax
\fill[EvalGray!10] (axis cs:\OldWindowLeft,0) rectangle (axis cs:\OldWindowRight,#3);
\fi
\fill[EvalFocus!8] (axis cs:\NewWindowLeft,0) rectangle (axis cs:\NewWindowRight,#3);
\draw[densely dashed,SCOAIMuted!65] (axis cs:.5,0)--(axis cs:.5,#3);
\ifnum#6>0\relax
\draw[CaseReference,densely dotted] (axis cs:#6,0)--(axis cs:#6,#3);
\node[anchor=north east,font=\scriptsize,text=CaseReference] at (axis cs:#6,#3) {Occasion};
\fi
\addplot[SCOAIInk,line width=.9pt] table[x=lead,y=observed]{\CaseTable};
\addplot[EvalGray,line width=1pt,dash pattern=on 4pt off 1.4pt on 1pt off 1.4pt] table[x=lead,y=baseline]{\CaseTable};
\addplot[EvalFocus,line width=1.25pt,dashed] table[x=lead,y=conditioned]{\CaseTable};
\legend{\KnowledgeOutcomeLabel,Original context,Updated context}
\end{axis}
\end{tikzpicture}\par\vspace{5pt}}

\newcommand{\KnowledgeFirstCase}{%
\begin{figure}[t]
\KnowledgeCasesIntro
\centering
\begin{evalcase}{K1}{Adding a previously missing back-to-school plan}
{\footnotesize\textbf{Background.} A back-to-school-sensitive product has a broadly flat pre-season sales trend despite daily fluctuations. The original context does not yet identify the approaching back-to-school period.\par}
\begin{knowledgebox}
\textbf{User update.} Include this product in the back-to-school campaign: promotional lead-in $T+1$--$T+6$, main activity $T+7$--$T+13$, and follow-up $T+14$--$T+20$. The numerical history and product attributes stay fixed.
\end{knowledgebox}
\KnowledgeCasePlot[4.1cm]{case_k1}{750}{7}{13}{0}{0}{0}
{\footnotesize\textbf{Response.} The updated forecast sustains the campaign lift and then eases. Both forecasts are smoother than the observed sales, and the updated trajectory retains a lower level and imperfect peak timing.\par}
\end{evalcase}
\caption{Online forecasting case with an added campaign plan. Shading marks the main activity; forecast revisions affect the sustained level without reproducing daily peaks.}
\label{fig:case_school}
\end{figure}
}
\newcommand{\KnowledgeOtherCases}{%
\begin{figure}[t]\centering
\begin{evalcase}{K2}{Moving a moderate forecast lift to a revised activity window}
{\footnotesize\textbf{Background.} A common campaign moves to a later period for an intermittently selling product. Numerical inputs and product scope stay fixed.\par}
\begin{knowledgebox}
\textbf{Earlier note.} Concentrate promotion during $T+7$--$T+12$.\par
\textbf{Updated note.} Move concentrated promotion to $T+14$--$T+19$, with regular product display beforehand and the usual rhythm afterward.
\end{knowledgebox}
\KnowledgeCasePlot[4.1cm]{case_k2}{1000}{14}{19}{0}{7}{12}
{\footnotesize\textbf{Response.} Forecast quantity shifts to the later window with a smooth, broad rise. Observed sales remain positive during the high-demand period and fluctuate irregularly. The predicted peak precedes the observed peak and remains lower.\par}
\end{evalcase}
\caption{Online forecasting case with a revised campaign schedule. Gray and orange shading mark the earlier and revised periods; level and timing differences remain visible.}
\label{fig:case_revision}
\end{figure}

\begin{figure}[t]\centering
\begin{evalcase}{K3}{An early shoulder and an aligned pre-occasion main peak}
{\footnotesize\textbf{Background.} An occasion-sensitive product is forecast around Father's Day. Both conditions share its $T+14$ date, the recent history, and the numerical business conditions.\par}
\begin{knowledgebox}
\textbf{User knowledge.} Father's Day is on $T+14$ in this calendar. Such products may build demand before the occasion and ease afterward; the expected lead-in is $T+6$--$T+14$.
\end{knowledgebox}
\KnowledgeCasePlot[4.1cm]{case_k3}{300}{6}{14}{14}{0}{0}
{\footnotesize\textbf{Response.} Observed sales form a broad occasion-related rise. The forecast has an earlier shoulder around $T+9$ and a main peak at $T+13$, aligned with the observed crest. The forecast underestimates the main peak and differs from the observed trajectory during the lead-in.\par}
\end{evalcase}
\caption{Online forecasting case with occasion knowledge. The main peaks align, with different amplitudes and an earlier forecast shoulder.}
\label{fig:case_fathers_day}
\end{figure}
}

\KnowledgeFirstCase
\KnowledgeOtherCases

\newcommand{\RegionalCasesIntro}{%
\subsection{Case Studies: National--DC Collaboration}
\label{sec:regional_cases}
Figures~\ref{fig:case_regional_decline}--\ref{fig:case_regional_selective} examine the selection and transfer of regional patterns. In the historical panels, green identifies applicable references and gray identifies histories with mismatched regional conditions; orange denotes the target. Pale daily histories and darker trailing seven-day means separate short-term fluctuations from the recent trend. Future panels compare observed sales, the blue comparison forecast, and the orange \model{} forecast. Each stream's retained historical mean defines index 100.
\par\medskip
}
\providecommand{\RegionalNationalColor}{CaseReference}
\providecommand{\RegionalPeerBColor}{CaseReference}
\providecommand{\RegionalPeerCColor}{CaseReference}
\definecolor{CaseControl}{HTML}{637FA2}
\newcommand{\CompactHistoryRow}[7]{%
\begin{axis}[case axis,name=#1,at={#2},anchor=north west,yshift=#3,
  scale only axis=true,width=.32\linewidth,height=.77cm,
  xmin=-20,xmax=0,ymax=#5,ytick={0,#5},xtick=\empty,#7,
  tick label style={font=\scriptsize,color=SCOAIMuted}]
\addplot[#6!45,line width=.5pt,forget plot]
 table[x expr=\thisrow{lead}+1,y=#4_history]{\CaseTable};
\addplot[#6,line width=1.05pt,forget plot]
 table[x expr=\thisrow{lead}+1,y=#4_ma7]{\CaseTable};
\node[anchor=east,font=\scriptsize\bfseries,text=#6,align=right]
 at (axis description cs:-.14,.5) {\csname HistoryLabel#4\endcsname};
\end{axis}}
\newcommand{\RegionalCompactPlot}[7]{%
\ReadCaseTable{data/#1_plot.csv}%
\pgfmathsetmacro{\FutureMid}{#4/2}%
\begin{tikzpicture}
\CompactHistoryRow{rhN}{(0,0)}{0cm}{national}{#2}{\RegionalNationalColor}
 {title={Historical inputs},title style={font=\scriptsize,color=SCOAISlate}}
\CompactHistoryRow{rhB}{(rhN.south west)}{-.24cm}{peer_b}{#2}{\RegionalPeerBColor}{} 
\CompactHistoryRow{rhC}{(rhB.south west)}{-.24cm}{peer_c}{#2}{\RegionalPeerCColor}{}
\CompactHistoryRow{rhA}{(rhC.south west)}{-.24cm}{focal}{#3}{EvalFocus}
 {xtick={-20,-10,0},xticklabels={$T-20$,$T-10$,$T$}}
\begin{axis}[case axis,at={(rhN.north east)},anchor=north west,
  xshift=1.15cm,scale only axis=true,width=.43\linewidth,height=3.60cm,
  xmin=1,xmax=28,ymax=#4,ytick={0,\FutureMid,#4},
  xtick={1,14,28},xticklabels={$1$,$H/2$,$H$},xlabel={Position in forecast window},
  title={Target future},
  title style={font=\scriptsize,color=EvalFocus},
  tick label style={font=\scriptsize,color=SCOAIMuted},
  legend style={font=\scriptsize,draw=none,fill=none,
    at={(-.15,-.25)},anchor=north,legend columns=3,column sep=4pt}]
\addplot[SCOAIInk,line width=.85pt]
 table[x expr=\thisrow{lead}+1,y=focal_observed]{\CaseTable};
\addplot[CaseControl,line width=.9pt,dash pattern=on 4pt off 1.4pt on 1pt off 1.4pt]
 table[x expr=\thisrow{lead}+1,y=control_forecast]{\CaseTable};
\addplot[EvalFocus,line width=1.1pt,dashed]
 table[x expr=\thisrow{lead}+1,y=focal_forecast]{\CaseTable};
\legend{#7,#5,#6}
\end{axis}
\end{tikzpicture}\par\vspace{3pt}}

\newcommand{\RegionalFirstCase}{%
\begin{figure}[t]
\RegionalCasesIntro
\centering
\begin{evalcase}{R1}{A lower local planning quantity after a regional high}
{\footnotesize\textbf{Background.} The target finishes its history with a large isolated burst. National and peer histories show different recent movements; the task is to assess the local quantity over the next window.\par}
\begingroup
\expandafter\def\csname HistoryLabelnational\endcsname{National}
\expandafter\def\csname HistoryLabelpeer_b\endcsname{DC-B}
\expandafter\def\csname HistoryLabelpeer_c\endcsname{DC-C}
\expandafter\def\csname HistoryLabelfocal\endcsname{Target}
\RegionalCompactPlot{case_r1}{600}{1000}{900}
 {No regional messages}{TimeHF~2.0}{Observed sales}
\endgroup
{\footnotesize\textbf{Response.} TimeHF~2.0 predicts a lower mid-window level than the no-regional-message model. Realized sales fall in the middle of the horizon and later rebound. The resulting total is closer, while the late burst remains underestimated.\par}
\end{evalcase}
\caption{Online local-quantity comparison; the late rebound remains underestimated.}
\label{fig:case_regional_decline}
\end{figure}
}
\newcommand{\RegionalOtherCases}{%
\begin{figure}[t]\centering
\begin{evalcase}{R2}{A restrained regional lift for sparse local demand}
{\footnotesize\textbf{Background.} Sales of a warm-season, weather-sensitive product rise in national and southern histories, while the northern target remains intermittent. The local weather outlook supports some later warming, with a weaker local response.\par}
\begingroup
\expandafter\def\csname HistoryLabelnational\endcsname{National}
\expandafter\def\csname HistoryLabelpeer_b\endcsname{South B}
\expandafter\def\csname HistoryLabelpeer_c\endcsname{South C}
\expandafter\def\csname HistoryLabelfocal\endcsname{North A}
\RegionalCompactPlot{case_r2}{500}{700}{500}
 {Single-series}{TimeHF~2.0}{Observed sales}
\endgroup
{\footnotesize\textbf{Response.} The forecast has a rising centerline with gentle pullbacks, above the single-series level but weaker than regional growth. Observed purchases become denser and their weekly level increases, with low and zero-sales days still present.\par}
\end{evalcase}
\caption{Online forecasting case with a locally anchored forecast and a restrained regional lift.}
\label{fig:case_regional_sparse}
\end{figure}

\begin{figure}[t]\centering
\begin{evalcase}{R3}{Rejecting unsuitable regional growth signals}
{\footnotesize\textbf{Background.} National and North B histories rise under cold conditions, while South A and the mild-climate peer South C retain a low, stable level. The national and northern growth patterns are unsuitable references for the southern target; South C provides the applicable regional context.\par}
\begingroup
\expandafter\def\csname HistoryLabelnational\endcsname{National}
\expandafter\def\csname HistoryLabelpeer_b\endcsname{North B}
\expandafter\def\csname HistoryLabelpeer_c\endcsname{South C}
\expandafter\def\csname HistoryLabelfocal\endcsname{South A}
\renewcommand{\RegionalNationalColor}{EvalGray}
\renewcommand{\RegionalPeerBColor}{EvalGray}
\renewcommand{\RegionalPeerCColor}{CaseReference}
\RegionalCompactPlot{case_r3}{350}{200}{220}
 {Single-series}{TimeHF~2.0}{Observed sales}
\endgroup
{\footnotesize\textbf{Response.} TimeHF~2.0 retains the target's local level instead of following the national and North B increases. Its trajectory remains consistent with the target and South C histories and avoids an unwarranted sales ramp. This behavior is consistent with applicability-aware regional sharing that transfers relevant patterns while suppressing mismatched growth signals.\par}
\end{evalcase}
\caption{Selective regional collaboration. The forecast retains the local sales level while avoiding the growth pattern of unsuitable national and northern references; South C supplies the applicable regional context.}
\label{fig:case_regional_selective}
\end{figure}
}

\RegionalFirstCase
\RegionalOtherCases

\Needspace{8\baselineskip}
\section{Discussion: From TimeHF~2.0 to TimeHF~3.0}
\label{sec:discussion}
TimeHF~2.0 establishes a business-conditioned forecasting interface: regional histories, structured conditions, and users' plans or experience enter the same numerical prediction process. Its practical contribution is to make flexible business descriptions usable alongside time-series evidence while retaining target-specific daily forecasts and planning-window quantities. The online comparisons demonstrate gains in both daily prediction and planning-window quantities. The loss comparison connects ordered-bin learning to robust planning quantities, and the product-type analysis examines shared/routed processing together with regional evidence. The trajectory cases connect these capabilities to campaign revisions, local sales regimes, and the selective use of regional evidence.

Sparse histories, stock-constrained observations, ambiguous plans, and abrupt local peaks still affect what can be learned. The missed rebound in Figure~\ref{fig:case_regional_decline} illustrates that a close accumulated quantity can coexist with substantial local errors. Useful daily timing, planning-window quantities, regional differences, and origin-time availability therefore remain essential when extending the forecasting interface.

Latent auto-encoder methods study explainable probabilistic forecasting in supply chains~\citep{hu2025explainable}. When a plan changes, a business user needs to understand which conditions matter, which products and dates are affected, and how the forecast has changed. Generative planning frameworks connect business intent with structured analytical workflows through language-model agents~\citep{yin2026generativeplanning}. TimeHF~2.0 produces sales distributions and quantities. Native explanatory output is therefore a further modeling objective: regional messages, attention patterns, and expert activations can support diagnosis, but do not themselves provide a faithful account of a forecast revision.

TimeHF~3.0 is motivated by this need for a more complete interaction with forecasts. The direction in Figure~\ref{fig:timehf_roadmap} is native multimodal-to-multimodal modeling: heterogeneous numerical and business inputs would support both forecasts and explanations within a common learned task representation. Product materials, plans, calendars, charts, and time series can carry complementary information. Forecast and explanatory outputs should refer to the same target, forecast origin, applicable conditions, and numerical evidence, allowing users to inspect a prediction and revise its context through one coherent interface.

Developing native explanations consequently requires evaluation beyond linguistic fluency. An explanation should remain consistent with the forecast values and supplied conditions, reflect uncertainty and residual disagreement, identify its temporal and regional scope, and respond coherently to controlled input changes. Forecast accuracy, explanation grounding, robustness to incomplete or conflicting context, and operational usability need to be assessed together. Forecast-revision explanations describe the model's conditional response; identifying the effect of actually executing a business plan additionally requires an appropriate decision evaluation. TimeHF~3.0 remains a research direction, with the numerical and semantic interfaces of TimeHF~2.0 providing the starting point.

\section{Conclusion}
\label{sec:conclusion}
\model{} connects regional histories, structured conditions, and flexible business knowledge within a common retail forecasting computation. Coverage-aware exchange and semantic alignment condition target-specific future queries, while shared/routed experts adapt processing across product and regional regimes. Ordered-bin supervision, a point-regression penalty, and cumulative learning connect robust daily prediction to operational planning quantities. Online parallel evaluation reports relative accuracy improvements of 18.25\% and 17.94\% for national daily and planning-window forecasts, and 19.44\% and 18.99\% for the corresponding DC forecasts, over the original production deep learning baseline. Component and loss comparisons support complementary information and supervision paths. Product-type results show gains across all four strata, with particularly strong DC improvements for seasonal and long-tail products. These forecasting capabilities motivate TimeHF~3.0: native multimodal prediction and explanation grounded in the same numerical evidence and business context.

\appendix
\Needspace{5cm}
\section{Supplementary Method Definitions}
\label{app:method_details}
\setcounter{equation}{0}
\renewcommand{\theequation}{A.\arabic{equation}}

\subsection{Numerical scales and ordered bins}
\label{app:normalization}
Each numerical stream uses its retained observations for its own statistics. For sales target $i$, $\mu_i$ is the retained-history mean and $\sigma_i$ is its standard deviation floored by a configured $s_{\min}>0$; the transform is $z_{i,t}=\asinh((y_{i,t}-\mu_i)/\sigma_i)$. Empty histories use an explicit missing-history indicator and a positive-scale prior from training information, with configured category/region and global fallbacks. Numerical covariates do not derive their scales from the current group's targets. Missingness, padding, origin-time availability, and supervision validity are distinct masks; an observed zero remains valid.

\label{app:bins}
For $K$ ordered bins, standard-normal reference quantiles define edges and finite centers:
\begin{equation}
\begin{aligned}
b_0&=-\infty,\quad b_K=+\infty,\quad b_k=\Phi^{-1}(k/K)\quad(1\leq k<K),\\
\nu_k&=\frac{\phi(b_{k-1})-\phi(b_k)}{\Phi(b_k)-\Phi(b_{k-1})}.
\end{aligned}
\label{eq:bins}
\end{equation}
Here $\Phi$ and $\phi$ are the standard-normal CDF and density. Equiprobability holds under this reference, not necessarily under empirical sales. Finite centers limit the point-decoding range conditional on history scale; text conditioning does not remove that limit. Equation~\ref{eq:decode} averages restored sales representatives, not transformed-space centers. Daily marginal quantiles cannot be summed into window quantiles. Fine central resolution supports routine variation, while wider tails represent broad extreme movements. Extreme values in the same terminal bin share a categorical label. The numerical and window penalties retain quantity constraints, and all valid sales observations remain in supervision and evaluation. Window supervision retains sensitivity to accumulated quantity across routine and extreme periods.

\subsection{Directed attention and text-connection conventions}
\label{app:attention}
\label{app:text}
Equation~\ref{eq:encoding} displays the temporal and FFN residuals explicitly and groups the directed variable-attention controls into the two messages. Within a variable-attention layer, regional and covariate messages share Q/K/V and output projections, read the same temporal states $U$, and have separate source sets and softmax denominators. Each set includes a null source of fixed score zero and value zero. A regional score combines content, type/coverage compatibility, log source reliability $\rho$, log pairwise applicability $\kappa$, and an allowed-edge mask. Covariate edges follow role, scope, and availability without these regional reliability factors. A covariate reader has no target-source branch.

The controlled target update underlying the two messages is
\begin{equation}
V_i=U_i+\gamma_i^Y\zeta_i^Y\mathcal B_i(W_O m_i^Y)
          +\gamma_i^C\zeta_i^C\mathcal B_i(W_O m_i^C),
\qquad
\mathcal B_i(x)=\frac{\operatorname{stopgrad}(\RMS(U_i))x}{\RMS(x)+\epsilon}.
\label{eq:var_update}
\end{equation}
Receiver gates $\gamma$ control use of each message; $\zeta$ is the real-source attention mass, averaged across heads. Its placement outside $\mathcal B$ preserves absolute attenuation. Attention controls direct forward messages rather than asserting causal independence. Coverage uses origin-visible service-city snapshots; empty coverage uses an unknown-region representation, and pooled city weights are contextual rather than identified sales shares.

The text connection projects language-model hidden representations into the forecasting dimension and uses attention-based alignment with numerical memory. The base unit includes product/region attributes and a short retained historical segment ending at $T$, with summaries generated after masking. Plan, event, and business-experience units use origin-visible descriptions and declared product/region scope and response ranges, including predefined leads/lags when applicable. User expectations are recorded as supplied qualitative beliefs rather than future observed outcomes. Event dates and response intervals are distinguished: the latter determine applicability, including lead-in or relaxation periods. Calendar-relative expressions use the forecast calendar in unit construction; this convention does not add a separate language-generation or reasoning module. Local events enter a national target only through an explicit national applicability relation. Separate descriptions share the alignment module and contribute to context for their applicable targets and dates.

The internal and query gates regulate how grounded text context enters forecasting. Invalid memory positions are excluded; a fully masked read returns a zero residual including output bias. Text padding is excluded from pooling, empty units are omitted, and an empty applicable set yields zero context. Disabling the language branch sets its context to zero; absence of events can still leave the base unit. Forecast supervision trains the semantic connection and its use in numerical prediction. Expert routing receives the decoder state after text-conditioned query construction and numerical-memory attention, so the semantic context is already available for expert selection.

\subsection{Forecast losses and target-level balance}
\label{app:learning}
\label{app:routing}
For the true ordered bin $b^\star$ and cumulative prediction $F_k=\sum_{j=1}^{k}p_j$, $\mathcal L_{\mathrm{CE}}=-\log p_{b^\star}$ and
\begin{equation}
\mathcal L_{\mathrm{RPS}}=\frac{1}{K-1}\sum_{k=1}^{K-1}
\left(F_k-\ind[b^\star\leq k]\right)^2.
\label{eq:ordinal}
\end{equation}
The daily regression penalty is $\mathcal L_{\mathrm{reg},i}=\operatorname{Mean}_{h\in\mathcal V_i}\mathcal R(\widehat y_{i,h},y_{i,h})$, where $\mathcal V_i$ contains valid daily labels and $\mathcal R$ denotes the configured point-regression penalty. It acts on the decoded quantity from Equation~\ref{eq:decode}; it is distinct from ordinal supervision and from the cumulative term. Fully labeled planning intervals within the forecast horizon enter Equation~\ref{eq:learning}. If no interval is eligible, the target contributes to the valid daily terms with a zero window term. The Huber function is $r^2/2$ for $|r|\leq\delta$ and $\delta(|r|-\delta/2)$ otherwise. Window totals refer to daily point forecasts, with no national/DC equality imposed.

For groups with valid labels at both levels, the national/DC aggregation and complete training objective are
\begin{equation}
\begin{aligned}
\mathcal L_g&=\alpha\mathcal L_{g,\mathrm{nat}}
 +(1-\alpha)\operatorname{Mean}_{i\in\mathcal D_g^{\mathrm{valid}}}\mathcal L_{g,i},\\
\mathcal L_{\mathrm{total}}&=\operatorname{Mean}_{g}\mathcal L_g
 +\lambda_{\mathrm{bal}}\mathcal L_{\mathrm{bal}}.
\end{aligned}
\label{eq:totalloss}
\end{equation}
When only one level is valid it receives the group's supervision weight; empty groups are skipped. Covariates and padding have no sales labels. An expert-balancing auxiliary loss~\citep{fedus2022switch} encourages balanced routed-expert use through selection frequencies and router probabilities, excluding shared processing and padding. Forecast and auxiliary coefficients are configuration parameters.

\bibliographystyle{plainnat}
\begingroup\fontsize{8.5}{10}\selectfont\setlength{\bibsep}{1pt}\interlinepenalty=10000\raggedright\bibliography{references}\endgroup

\begin{thebibliography}{47}
\providecommand{\natexlab}[1]{#1}
\providecommand{\url}[1]{\texttt{#1}}
\expandafter\ifx\csname urlstyle\endcsname\relax
  \providecommand{\doi}[1]{doi: #1}\else
  \providecommand{\doi}{doi: \begingroup \urlstyle{rm}\Url}\fi

\bibitem[Ansari et~al.(2024)Ansari, Stella, Turkmen, et~al.]{ansari2024chronos}
Abdul~Fatir Ansari, Lorenzo Stella, Caner Turkmen, et~al.
\newblock Chronos: Learning the language of time series, 2024.
\newblock URL \url{https://arxiv.org/abs/2403.07815}.

\bibitem[Ansari et~al.(2025)Ansari, Shchur, K{\"u}ken, Auer, Han, Mercado,
  Rangapuram, Shen, Stella, Zhang, Goswami, Kapoor, Maddix, Guerron, Hu, Yin,
  Erickson, Desai, Wang, Rangwala, Karypis, Wang, and
  Bohlke-Schneider]{ansari2025chronos2}
Abdul~Fatir Ansari, Oleksandr Shchur, Jaris K{\"u}ken, Andreas Auer, Boran Han,
  Pedro Mercado, Syama~Sundar Rangapuram, Huibin Shen, Lorenzo Stella, Xiyuan
  Zhang, Mononito Goswami, Shubham Kapoor, Danielle~C. Maddix, Pablo Guerron,
  Tony Hu, Junming Yin, Nick Erickson, Prateek~Mutalik Desai, Hao Wang, Huzefa
  Rangwala, George Karypis, Yuyang Wang, and Michael Bohlke-Schneider.
\newblock {Chronos-2}: From univariate to universal forecasting.
\newblock arXiv:2510.15821, 2025.
\newblock URL \url{https://arxiv.org/abs/2510.15821}.

\bibitem[Athanasopoulos et~al.(2017)Athanasopoulos, Hyndman, Kourentzes, and
  Petropoulos]{athanasopoulos2017temporal}
George Athanasopoulos, Rob~J. Hyndman, Nikolaos Kourentzes, and Fotios
  Petropoulos.
\newblock Forecasting with temporal hierarchies.
\newblock \emph{European Journal of Operational Research}, 262\penalty0
  (1):\penalty0 60--74, 2017.
\newblock \doi{10.1016/j.ejor.2017.02.046}.
\newblock URL \url{https://robjhyndman.com/publications/temporal-hierarchies/}.

\bibitem[Auer et~al.(2025)Auer, Podest, Klotz, B{\"o}ck, Klambauer, and
  Hochreiter]{auer2025tirex}
Andreas Auer, Patrick Podest, Daniel Klotz, Sebastian B{\"o}ck, G{\"u}nter
  Klambauer, and Sepp Hochreiter.
\newblock {TiRex}: Zero-shot forecasting across long and short horizons with
  enhanced in-context learning.
\newblock arXiv:2505.23719, 2025.
\newblock URL \url{https://arxiv.org/abs/2505.23719}.

\bibitem[Das et~al.(2024)Das, Kong, Sen, and Zhou]{das2024timesfm}
Abhimanyu Das, Weihao Kong, Rajat Sen, and Yichen Zhou.
\newblock A decoder-only foundation model for time-series forecasting.
\newblock In \emph{Proceedings of the 41st International Conference on Machine
  Learning}, volume 235 of \emph{Proceedings of Machine Learning Research},
  pages 10148--10167, 2024.
\newblock URL \url{https://proceedings.mlr.press/v235/das24c.html}.

\bibitem[Fedus et~al.(2022)Fedus, Zoph, and Shazeer]{fedus2022switch}
William Fedus, Barret Zoph, and Noam Shazeer.
\newblock {Switch Transformers}: Scaling to trillion parameter models with
  simple and efficient sparsity.
\newblock \emph{Journal of Machine Learning Research}, 23\penalty0
  (120):\penalty0 1--39, 2022.
\newblock URL \url{https://www.jmlr.org/papers/v23/21-0998.html}.

\bibitem[Gneiting and Raftery(2007)]{gneiting2007scoring}
Tilmann Gneiting and Adrian~E. Raftery.
\newblock Strictly proper scoring rules, prediction, and estimation.
\newblock \emph{Journal of the American Statistical Association}, 102\penalty0
  (477):\penalty0 359--378, 2007.
\newblock \doi{10.1198/016214506000001437}.

\bibitem[Gruver et~al.(2023)Gruver, Finzi, Qiu, and Wilson]{gruver2023llmtime}
Nate Gruver, Marc Finzi, Shikai Qiu, and Andrew~Gordon Wilson.
\newblock Large language models are zero-shot time series forecasters.
\newblock In \emph{Advances in Neural Information Processing Systems},
  volume~36, 2023.
\newblock URL
  \url{https://papers.neurips.cc/paper_files/paper/2023/hash/3eb7ca52e8207697361b2c0fb3926511-Abstract-Conference.html}.

\bibitem[Han et~al.(2023{\natexlab{a}})Han, Chen, Zheng, Liu, Wang, Cheng, and
  Li]{han2023dominantsets}
Zhongxuan Han, Chaochao Chen, Xiaolin Zheng, Weiming Liu, Jun Wang, Wenjie
  Cheng, and Yuyuan Li.
\newblock In-processing user constrained dominant sets for user-oriented
  fairness in recommender systems.
\newblock In \emph{Proceedings of the 31st ACM International Conference on
  Multimedia}, pages 6190--6201, 2023{\natexlab{a}}.
\newblock \doi{10.1145/3581783.3613831}.
\newblock URL \url{https://doi.org/10.1145/3581783.3613831}.

\bibitem[Han et~al.(2023{\natexlab{b}})Han, Zheng, Chen, Cheng, and
  Yao]{han2023crossdomain}
Zhongxuan Han, Xiaolin Zheng, Chaochao Chen, Wenjie Cheng, and Yang Yao.
\newblock Intra and inter domain hypergraph convolutional network for
  cross-domain recommendation.
\newblock In \emph{Proceedings of the ACM Web Conference 2023}, pages 449--459,
  2023{\natexlab{b}}.
\newblock \doi{10.1145/3543507.3583402}.
\newblock URL \url{https://doi.org/10.1145/3543507.3583402}.

\bibitem[Han et~al.(2024)Han, Chen, Zheng, Li, Liu, Yao, Li, and
  Yin]{han2024transport}
Zhongxuan Han, Chaochao Chen, Xiaolin Zheng, Meng Li, Weiming Liu, Binhui Yao,
  Yuyuan Li, and Jianwei Yin.
\newblock Intra- and inter-group optimal transport for user-oriented fairness
  in recommender systems.
\newblock \emph{Proceedings of the AAAI Conference on Artificial Intelligence},
  38\penalty0 (8):\penalty0 8463--8471, 2024.
\newblock \doi{10.1609/aaai.v38i8.28689}.
\newblock URL \url{https://ojs.aaai.org/index.php/AAAI/article/view/28689}.

\bibitem[Hu et~al.(2024)Hu, Qi, Lee, Shen, Liu, Zhu, and
  Kang]{hu2024supercharged}
Hao Hu, Yongzhi Qi, Hau~L. Lee, Zuo-Jun~Max Shen, Curtis Liu, Weimeng Zhu, and
  Ningxuan Kang.
\newblock Supercharged by advanced analytics, {JD.com} attains agility,
  resilience, and shared value across its supply chain.
\newblock \emph{INFORMS Journal on Applied Analytics}, 54\penalty0
  (1):\penalty0 54--70, 2024.
\newblock \doi{10.1287/inte.2023.0078}.
\newblock URL \url{https://pubsonline.informs.org/doi/10.1287/inte.2023.0078}.

\bibitem[Hu et~al.(2025)Hu, Wu, Shi, Qi, Zhang, Han, Kang, Chen, and
  Wang]{hu2025explainable}
Hao Hu, Yuting Wu, Zhengxin Shi, Yongzhi Qi, Jianshen Zhang, Shuyu Han,
  Ningxuan Kang, Jian Chen, and Ling Wang.
\newblock Explainable probabilistic forecasting for time series in supply
  chains: A latent auto-encoder approach.
\newblock \emph{International Journal of Production Research}, 63\penalty0
  (24):\penalty0 9933--9953, 2025.
\newblock \doi{10.1080/00207543.2025.2541860}.
\newblock URL
  \url{https://www.tandfonline.com/doi/full/10.1080/00207543.2025.2541860}.

\bibitem[Huber(1964)]{huber1964robust}
Peter~J. Huber.
\newblock Robust estimation of a location parameter.
\newblock \emph{The Annals of Mathematical Statistics}, 35\penalty0
  (1):\penalty0 73--101, 1964.
\newblock \doi{10.1214/aoms/1177703732}.
\newblock URL \url{https://doi.org/10.1214/aoms/1177703732}.

\bibitem[Hyndman et~al.(2011)Hyndman, Ahmed, Athanasopoulos, and
  Shang]{hyndman2011hierarchical}
Rob~J. Hyndman, Roman~A. Ahmed, George Athanasopoulos, and Han~Lin Shang.
\newblock Optimal combination forecasts for hierarchical time series.
\newblock \emph{Computational Statistics \& Data Analysis}, 55\penalty0
  (9):\penalty0 2579--2589, 2011.
\newblock \doi{10.1016/j.csda.2011.03.006}.
\newblock URL \url{https://robjhyndman.com/publications/hierarchical/}.

\bibitem[Jin et~al.(2024)Jin, Wang, Ma, Chu, Zhang, Shi, Chen, Liang, Li, Pan,
  and Wen]{jin2024timellm}
Ming Jin, Shiyu Wang, Lintao Ma, Zhixuan Chu, James~Y. Zhang, Xiaoming Shi,
  Pin-Yu Chen, Yuxuan Liang, Yuan-Fang Li, Shirui Pan, and Qingsong Wen.
\newblock {Time-LLM}: Time series forecasting by reprogramming large language
  models.
\newblock In \emph{The Twelfth International Conference on Learning
  Representations}, 2024.
\newblock URL \url{https://arxiv.org/abs/2310.01728}.

\bibitem[Khwaja et~al.(2026)Khwaja, Lettieri, Woo, et~al.]{khwaja2026toto2}
Emaad Khwaja, Chris Lettieri, Gerald Woo, et~al.
\newblock {Toto 2.0}: Time series forecasting enters the scaling era.
\newblock arXiv:2605.20119, 2026.
\newblock URL \url{https://arxiv.org/abs/2605.20119}.

\bibitem[Lei et~al.(2023)Lei, Hu, Geng, Zhang, Qi, Liu, and
  Shen]{lei2023lifecycle}
Dazhou Lei, Hao Hu, Dongyang Geng, Jianshen Zhang, Yongzhi Qi, Sheng Liu, and
  Zuo-Jun~Max Shen.
\newblock New product life cycle curve modeling and forecasting with product
  attributes and promotion: A {Bayesian} functional approach.
\newblock \emph{Production and Operations Management}, 32\penalty0
  (2):\penalty0 655--673, 2023.
\newblock \doi{10.1111/poms.13892}.
\newblock URL \url{https://journals.sagepub.com/doi/10.1111/poms.13892}.

\bibitem[Lei et~al.(2025)Lei, Qi, Liu, Geng, Zhang, Hu, and
  Shen]{lei2025pooling}
Dazhou Lei, Yongzhi Qi, Sheng Liu, Dongyang Geng, Jianshen Zhang, Hao Hu, and
  Zuo-Jun~Max Shen.
\newblock Pooling and boosting for demand prediction in retail: A transfer
  learning approach.
\newblock \emph{Manufacturing \& Service Operations Management}, 27\penalty0
  (6):\penalty0 1779--1794, 2025.
\newblock \doi{10.1287/msom.2022.0453}.
\newblock URL \url{https://pubsonline.informs.org/doi/10.1287/msom.2022.0453}.

\bibitem[Lim et~al.(2021)Lim, Arik, Loeff, and Pfister]{lim2021tft}
Bryan Lim, Sercan~O. Arik, Nicolas Loeff, and Tomas Pfister.
\newblock Temporal fusion transformers for interpretable multi-horizon time
  series forecasting.
\newblock \emph{International Journal of Forecasting}, 37\penalty0
  (4):\penalty0 1748--1764, 2021.
\newblock \doi{10.1016/j.ijforecast.2021.03.012}.
\newblock URL \url{https://doi.org/10.1016/j.ijforecast.2021.03.012}.

\bibitem[Lin et~al.(2021)Lin, Zhang, Zhang, Qi, and Hu]{lin2021seasonal}
Xiaoming Lin, Ben Zhang, Jianshen Zhang, Yongzhi Qi, and Hao Hu.
\newblock A practical framework for forecasting stock keeping unit level
  seasonal sales.
\newblock In \emph{2021 7th International Conference on Big Data and
  Information Analytics (BigDIA)}, pages 393--398, 2021.
\newblock \doi{10.1109/BigDIA53151.2021.9619627}.
\newblock URL \url{https://ieeexplore.ieee.org/document/9619627}.

\bibitem[Liu et~al.(2025)Liu, Xu, Miao, Yang, Zhang, Long, Li, and
  Zhao]{liu2025timecma}
Chenxi Liu, Qianxiong Xu, Hao Miao, Sun Yang, Lingzheng Zhang, Cheng Long,
  Ziyue Li, and Rui Zhao.
\newblock {TimeCMA}: Towards {LLM}-empowered multivariate time series
  forecasting via cross-modality alignment.
\newblock \emph{Proceedings of the AAAI Conference on Artificial Intelligence},
  39\penalty0 (18):\penalty0 18780--18788, 2025.
\newblock \doi{10.1609/aaai.v39i18.34067}.
\newblock URL \url{https://ojs.aaai.org/index.php/AAAI/article/view/34067}.

\bibitem[Liu et~al.(2024{\natexlab{a}})Liu, Xu, Zhao, Kong, Kamarthi, Sasanur,
  Sharma, Cui, Wen, Zhang, and Prakash]{liu2024timemmd}
Haoxin Liu, Shangqing Xu, Zhiyuan Zhao, Lingkai Kong, Harshavardhan Kamarthi,
  Aditya~B. Sasanur, Megha Sharma, Jiaming Cui, Qingsong Wen, Chao Zhang, and
  B.~Aditya Prakash.
\newblock {Time-MMD}: Multi-domain multimodal dataset for time series analysis.
\newblock In \emph{Advances in Neural Information Processing Systems},
  volume~37, 2024{\natexlab{a}}.
\newblock URL
  \url{https://proceedings.neurips.cc/paper_files/paper/2024/file/8e7768122f3eeec6d77cd2b424b72413-Paper-Datasets_and_Benchmarks_Track.pdf}.

\bibitem[Liu et~al.(2024{\natexlab{b}})Liu, Hu, Li, Diao, Liang, Hooi, and
  Zimmermann]{liu2024unitime}
Xu~Liu, Junfeng Hu, Yuan Li, Shizhe Diao, Yuxuan Liang, Bryan Hooi, and Roger
  Zimmermann.
\newblock {UniTime}: A language-empowered unified model for cross-domain time
  series forecasting.
\newblock In \emph{Proceedings of the ACM Web Conference 2024}, pages
  4095--4106, 2024{\natexlab{b}}.
\newblock \doi{10.1145/3589334.3645434}.
\newblock URL \url{https://arxiv.org/abs/2310.09751}.

\bibitem[Liu et~al.(2024{\natexlab{c}})Liu, Liu, Woo, Aksu, Liang, Zimmermann,
  Liu, Savarese, Xiong, and Sahoo]{liu2024moirai}
Xu~Liu, Juncheng Liu, Gerald Woo, Taha Aksu, Yuxuan Liang, Roger Zimmermann,
  Chenghao Liu, Silvio Savarese, Caiming Xiong, and Doyen Sahoo.
\newblock {Moirai-MoE}: Empowering time series foundation models with sparse
  mixture of experts.
\newblock arXiv:2410.10469, 2024{\natexlab{c}}.
\newblock URL \url{https://arxiv.org/abs/2410.10469}.

\bibitem[Liu et~al.(2026)Liu, Huang, Yu, Zhou, Wang, Yan, Zhou, Yao, Han, Li,
  Zhang, Chen, Qi, Qi, and Shen]{liu2026scoai}
Yang Liu, Yulin Huang, Xue Yu, Haowei Zhou, Jiaxin Wang, Qiqi Yan, Tianqi Zhou,
  Yi~Yao, Zhongxuan Han, Shaomin Li, Jianshen Zhang, Zhe Chen, Yongzhi Qi, Wei
  Qi, and Zuo-Jun~(Max) Shen.
\newblock {ScoAI}: Toward {OR}-native decision intelligence.
\newblock Preprints.org, 2026.
\newblock URL \url{https://doi.org/10.20944/preprints202610.0845.v1}.
\newblock Version 1, posted 10 October 2026; not peer-reviewed.

\bibitem[Liu et~al.(2024{\natexlab{d}})Liu, Hu, Zhang, Wu, Wang, Ma, and
  Long]{liu2024itransformer}
Yong Liu, Tengge Hu, Haoran Zhang, Haixu Wu, Shiyu Wang, Lintao Ma, and
  Mingsheng Long.
\newblock {iTransformer}: Inverted transformers are effective for time series
  forecasting.
\newblock In \emph{International Conference on Learning Representations},
  2024{\natexlab{d}}.
\newblock URL \url{https://arxiv.org/abs/2310.06625}.

\bibitem[Nie et~al.(2023)Nie, Nguyen, Sinthong, and
  Kalagnanam]{nie2023patchtst}
Yuqi Nie, Nam~H. Nguyen, Phanwadee Sinthong, and Jayant Kalagnanam.
\newblock A time series is worth 64 words: Long-term forecasting with
  transformers.
\newblock In \emph{International Conference on Learning Representations}, 2023.
\newblock URL \url{https://arxiv.org/abs/2211.14730}.

\bibitem[Oreshkin et~al.(2020)Oreshkin, Carpov, Chapados, and
  Bengio]{oreshkin2020nbeats}
Boris~N. Oreshkin, Dmitri Carpov, Nicolas Chapados, and Yoshua Bengio.
\newblock {N-BEATS}: Neural basis expansion analysis for interpretable time
  series forecasting.
\newblock In \emph{International Conference on Learning Representations}, 2020.
\newblock URL \url{https://arxiv.org/abs/1905.10437}.

\bibitem[Pineda~Arango et~al.(2025)Pineda~Arango, Mercado, Kapoor, Ansari,
  Stella, Shen, Senetaire, Turkmen, Shchur, Maddix, Bohlke-Schneider, Wang, and
  Rangapuram]{arango2025chronosx}
Sebastian Pineda~Arango, Pedro Mercado, Shubham Kapoor, Abdul~Fatir Ansari,
  Lorenzo Stella, Huibin Shen, Hugo Senetaire, Caner Turkmen, Oleksandr Shchur,
  Danielle~C. Maddix, Michael Bohlke-Schneider, Yuyang Wang, and Syama~Sundar
  Rangapuram.
\newblock {ChronosX}: Adapting pretrained time series models with exogenous
  variables.
\newblock arXiv:2503.12107, 2025.
\newblock URL \url{https://arxiv.org/abs/2503.12107}.

\bibitem[Qi et~al.(2023)Qi, Shi, Qi, Ma, Yuan, Wu, and Shen]{qi2023inventory}
Meng Qi, Yuanyuan Shi, Yongzhi Qi, Chenxin Ma, Rong Yuan, Di~Wu, and
  Zuo-Jun~Max Shen.
\newblock A practical end-to-end inventory management model with deep learning.
\newblock \emph{Management Science}, 69\penalty0 (2):\penalty0 759--773, 2023.
\newblock \doi{10.1287/mnsc.2022.4564}.
\newblock URL \url{https://pubsonline.informs.org/doi/10.1287/mnsc.2022.4564}.

\bibitem[Qi et~al.(2025)Qi, Hu, Lei, Zhang, Shi, Huang, Chen, Lin, and
  Shen]{qi2025timehf}
Yongzhi Qi, Hao Hu, Dazhou Lei, Jianshen Zhang, Zhengxin Shi, Yulin Huang,
  Zhengyu Chen, Xiaoming Lin, and Zuo-Jun~Max Shen.
\newblock {TimeHF}: Billion-scale time series models guided by human feedback.
\newblock arXiv:2501.15942, 2025.
\newblock URL \url{https://arxiv.org/abs/2501.15942}.

\bibitem[Rangapuram et~al.(2018)Rangapuram, Seeger, Gasthaus, Stella, Wang, and
  Januschowski]{rangapuram2018deepstate}
Syama~Sundar Rangapuram, Matthias~W. Seeger, Jan Gasthaus, Lorenzo Stella,
  Yuyang Wang, and Tim Januschowski.
\newblock Deep state space models for time series forecasting.
\newblock In \emph{Advances in Neural Information Processing Systems},
  volume~31, 2018.
\newblock URL
  \url{https://papers.nips.cc/paper_files/paper/2018/hash/5cf68969fb67aa6082363a6d4e6468e2-Abstract.html}.

\bibitem[Rangapuram et~al.(2021)Rangapuram, Werner, Benidis, Mercado, Gasthaus,
  and Januschowski]{rangapuram2021hierarchical}
Syama~Sundar Rangapuram, Lucien~D. Werner, Konstantinos Benidis, Pedro Mercado,
  Jan Gasthaus, and Tim Januschowski.
\newblock End-to-end learning of coherent probabilistic forecasts for
  hierarchical time series.
\newblock In \emph{Proceedings of the 38th International Conference on Machine
  Learning}, volume 139 of \emph{Proceedings of Machine Learning Research},
  pages 8832--8843, 2021.
\newblock URL \url{https://proceedings.mlr.press/v139/rangapuram21a.html}.

\bibitem[Salinas et~al.(2017)Salinas, Flunkert, and
  Gasthaus]{salinas2017deepar}
David Salinas, Valentin Flunkert, and Jan Gasthaus.
\newblock {DeepAR}: Probabilistic forecasting with autoregressive recurrent
  networks.
\newblock arXiv:1704.04110, 2017.
\newblock URL \url{https://arxiv.org/abs/1704.04110}.

\bibitem[Shazeer et~al.(2017)Shazeer, Mirhoseini, Maziarz, Davis, Le, Hinton,
  and Dean]{shazeer2017moe}
Noam Shazeer, Azalia Mirhoseini, Krzysztof Maziarz, Andy Davis, Quoc Le,
  Geoffrey Hinton, and Jeff Dean.
\newblock Outrageously large neural networks: The sparsely-gated
  mixture-of-experts layer.
\newblock In \emph{International Conference on Learning Representations}, 2017.
\newblock URL \url{https://arxiv.org/abs/1701.06538}.

\bibitem[Shen et~al.(2025)Shen, Sun, Qi, Hu, Kang, Zhang, Wang, and
  Lin]{shen2025fulfillment}
Zuo-Jun~Max Shen, Shuo Sun, Yongzhi Qi, Hao Hu, Ningxuan Kang, Jianshen Zhang,
  Xin Wang, and Xiaoming Lin.
\newblock {JD.com} improves fulfillment efficiency with data-driven integrated
  assortment planning and inventory allocation.
\newblock \emph{INFORMS Journal on Applied Analytics}, 55\penalty0
  (5):\penalty0 386--398, 2025.
\newblock \doi{10.1287/inte.2025.0245}.
\newblock URL \url{https://pubsonline.informs.org/doi/10.1287/inte.2025.0245}.

\bibitem[Tan et~al.(2024)Tan, Merrill, Gupta, Althoff, and
  Hartvigsen]{tan2024useful}
Mingtian Tan, Mike~A. Merrill, Vinayak Gupta, Tim Althoff, and Thomas
  Hartvigsen.
\newblock Are language models actually useful for time series forecasting?
\newblock In \emph{Advances in Neural Information Processing Systems},
  volume~37, 2024.
\newblock URL \url{https://arxiv.org/abs/2406.16964}.

\bibitem[Wang et~al.(2022)Wang, Deng, Shen, Hu, and Qi]{wang2022digitaltwin}
Lu~Wang, Tianhu Deng, Zuo-Jun~Max Shen, Hao Hu, and Yongzhi Qi.
\newblock Digital twin-driven smart supply chain.
\newblock \emph{Frontiers of Engineering Management}, 9\penalty0 (1):\penalty0
  56--70, 2022.
\newblock \doi{10.1007/s42524-021-0186-9}.
\newblock URL
  \url{https://journal.hep.com.cn/fem/EN/10.1007/s42524-021-0186-9}.

\bibitem[Wickramasuriya et~al.(2019)Wickramasuriya, Athanasopoulos, and
  Hyndman]{wickramasuriya2019mint}
Shanika~L. Wickramasuriya, George Athanasopoulos, and Rob~J. Hyndman.
\newblock Optimal forecast reconciliation for hierarchical and grouped time
  series through trace minimization.
\newblock \emph{Journal of the American Statistical Association}, 114\penalty0
  (526):\penalty0 804--819, 2019.
\newblock \doi{10.1080/01621459.2018.1448825}.
\newblock URL \url{https://robjhyndman.com/publications/mint/}.

\bibitem[Williams et~al.(2025)Williams, Ashok, Marcotte, Zantedeschi,
  Subramanian, Riachi, Requeima, Lacoste, Rish, Chapados, and
  Drouin]{williams2025context}
Andrew~Robert Williams, Arjun Ashok, {\'{E}}tienne Marcotte, Valentina
  Zantedeschi, Jithendaraa Subramanian, Roland Riachi, James Requeima,
  Alexandre Lacoste, Irina Rish, Nicolas Chapados, and Alexandre Drouin.
\newblock Context is key: A benchmark for forecasting with essential textual
  information.
\newblock In \emph{Proceedings of the 42nd International Conference on Machine
  Learning}, volume 267 of \emph{Proceedings of Machine Learning Research},
  pages 66887--66944, 2025.
\newblock URL \url{https://proceedings.mlr.press/v267/williams25a.html}.

\bibitem[Woo et~al.(2024)Woo, Liu, Kumar, Xiong, Savarese, and
  Sahoo]{woo2024moirai}
Gerald Woo, Chenghao Liu, Akshat Kumar, Caiming Xiong, Silvio Savarese, and
  Doyen Sahoo.
\newblock Unified training of universal time series forecasting transformers.
\newblock arXiv:2402.02592, 2024.
\newblock URL \url{https://arxiv.org/abs/2402.02592}.

\bibitem[Yao et~al.(2025)Yao, Li, Dai, Zhang, Gong, Wang, and
  Lv]{yao2025captime}
Yueyang Yao, Jiajun Li, Xingyuan Dai, Mengmeng Zhang, Xiaoyan Gong, Fei-Yue
  Wang, and Yisheng Lv.
\newblock Context-aware probabilistic modeling with {LLM} for multimodal time
  series forecasting.
\newblock arXiv:2505.10774, 2025.
\newblock URL \url{https://arxiv.org/abs/2505.10774}.

\bibitem[Yin et~al.(2026)Yin, Qi, Zhang, Geng, Chen, Hu, Qi, and
  Shen]{yin2026generativeplanning}
Jiaheng Yin, Yongzhi Qi, Jianshen Zhang, Dongyang Geng, Zhengyu Chen, Hao Hu,
  Wei Qi, and Zuo-Jun~Max Shen.
\newblock Rethinking supply chain planning: A generative paradigm.
\newblock arXiv:2509.03811v2, 2026.
\newblock URL \url{https://arxiv.org/abs/2509.03811v2}.

\bibitem[Zhang and Yan(2023)]{zhang2023crossformer}
Yunhao Zhang and Junchi Yan.
\newblock {Crossformer}: Transformer utilizing cross-dimension dependency for
  multivariate time series forecasting.
\newblock In \emph{International Conference on Learning Representations}, 2023.
\newblock URL \url{https://openreview.net/forum?id=vSVLM2j9eie}.

\bibitem[Zhao et~al.(2026)Zhao, Yu, Qi, Hu, Zhang, Ma, Han, Qi, and
  Shen]{zhao2026orpr}
Lingjie Zhao, Xue Yu, Yongzhi Qi, Hao Hu, Jianshen Zhang, Yingzheng Ma, Shuyu
  Han, Wei Qi, and Zuo-Jun~Max Shen.
\newblock {ORPR}: An {OR}-guided pretrain-then-reinforce learning model for
  inventory management.
\newblock arXiv:2512.19001v2, 2026.
\newblock URL \url{https://arxiv.org/abs/2512.19001v2}.

\bibitem[Zhou et~al.(2023)Zhou, Niu, Wang, Sun, and Jin]{zhou2023onefitsall}
Tian Zhou, Peisong Niu, Xue Wang, Liang Sun, and Rong Jin.
\newblock One fits all: Power general time series analysis by pretrained {LM}.
\newblock In \emph{Advances in Neural Information Processing Systems},
  volume~36, 2023.
\newblock URL \url{https://arxiv.org/abs/2302.11939}.

\end{thebibliography}
\end{document}